%% file: thesis.tex
\documentclass[a4paper,12pt,times,numbered,print,index,custombib]{Classes/PhDThesisPSnPDF}

\input{Preamble/preamble}

\input{thesis-info}

\begin{document}

\frontmatter

\maketitle

\include{Dedication/dedication}
\include{Abstract/abstract}
\include{Contributions/contributions}


\tableofcontents




\printnomenclature

\mainmatter

\include{Chapter1/chapter1}

\include{Chapter2/chapter2}
\include{Chapter3/chapter3}


\include{Conclusions/conclusions}


\begin{spacing}{0.9}


\bibliographystyle{apalike}
\cleardoublepage
\bibliography{References/references} 



\end{spacing}


\begin{appendices} 

\end{appendices}


\end{document}

%% file: thesis-info.tex
\title{Deep Learning Training Procedure Augmentations}

\subtitle{Finding the right path is just as important as having a powerful vehicle}

\author{Cristian Simionescu}

\dept{Faculty of Computer Science}

\university{Alexandru Ioan Cuza University}
\crest{\includegraphics[width=0.2\textwidth]{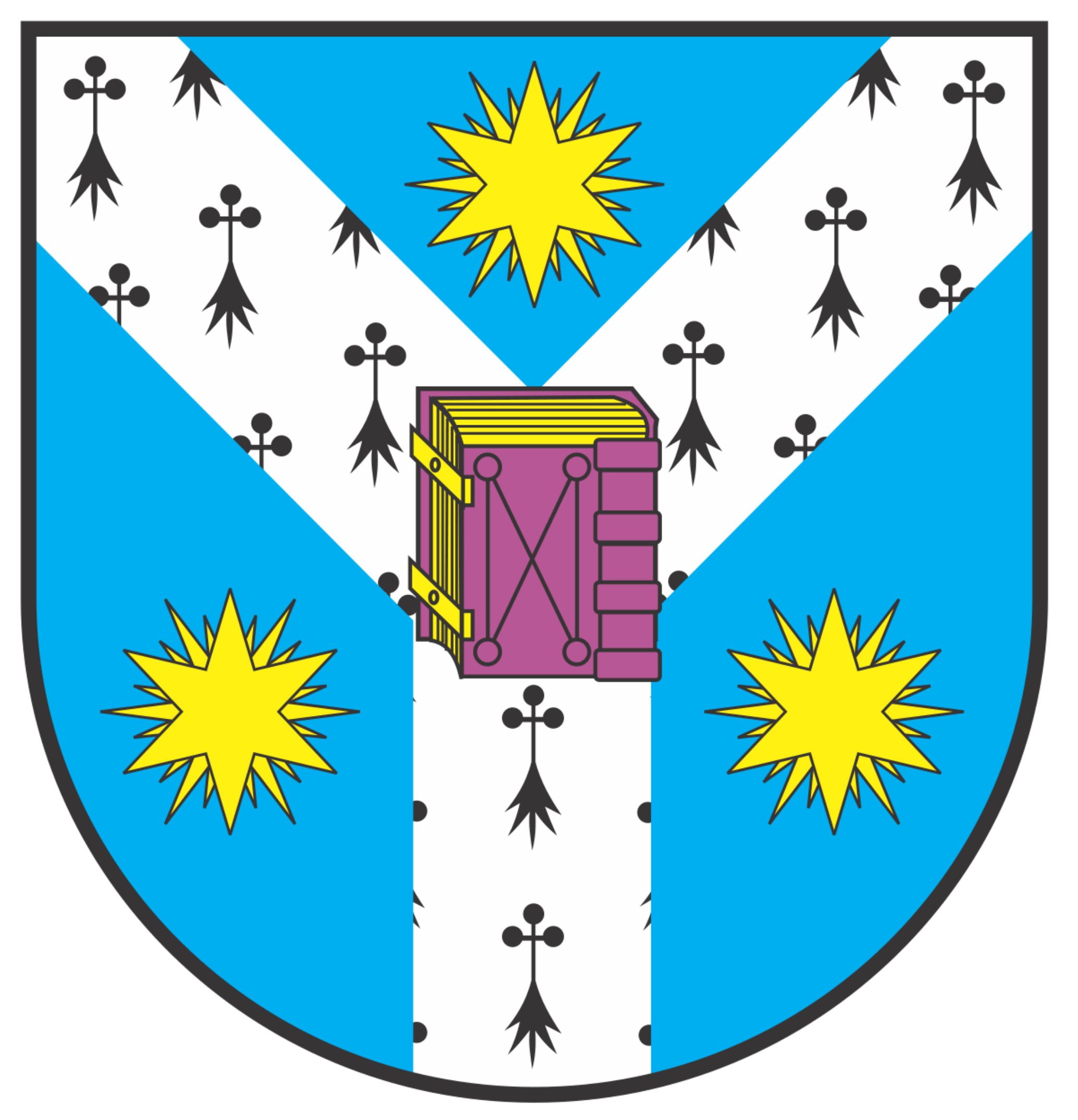}}

\collegeshield{\includegraphics[width=0.5\textwidth]{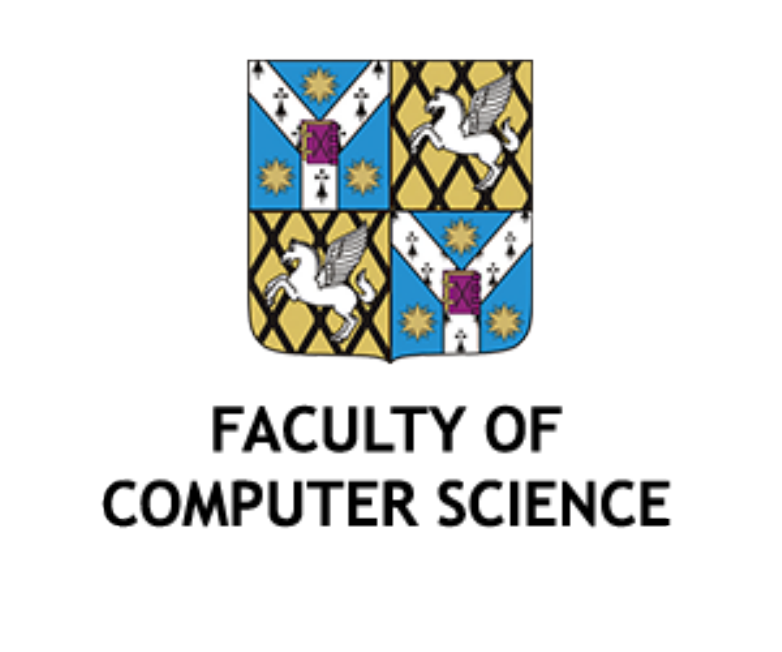}}

\supervisor{Associate Prof. Adrian Iftene}

\supervisorlinewidth{0.50\textwidth}

\degreetitle{Master of Science}

\college{Faculty of Computer Science}

\degreedate{July 2020} 

\subject{CAN}

%% file: Dedication/dedication.tex

\begin{dedication} 

This thesis is dedicated to my parents whose examples have always taught me only hard work will bring you closer to the things you aspire to achieve. I also dedicate this to Madalina who has always been a constant source of support and encouragement during the challenges of my university life. 

\end{dedication}

%% file: Abstract/abstract.tex
\begin{abstract}

Recent advances in Deep Learning have greatly improved performance on various tasks such as object detection, image segmentation, sentiment analysis. The focus of most research directions up until very recently has been on beating state-of-the-art results. This has materialized in the utilization of bigger and bigger models and techniques which help the training procedure to extract more predictive power out of a given dataset. While this has lead to great results, many of which with real-world applications, other relevant aspects of deep learning have remained neglected and unknown. In this work, we will present several novel deep learning training techniques which, while capable of offering significant performance gains they also reveal several interesting analysis results regarding convergence speed, optimization landscape smoothness, and adversarial robustness.
\newline
The methods presented in this work are the following:
\begin{enumerate}
    \item Perfect Ordering Approximation; a generalized model agnostic curriculum learning approach. The results show the effectiveness of the technique for improving training time as well as offer some new insight into the training process of deep networks.
    \item Cascading Sum Augmentation; an extension of mixup capable of utilizing more data points for linear interpolation by leveraging a smoother optimization landscape. This can be used for computer vision tasks in order to improve both prediction performance as well as improve passive model robustness.
\end{enumerate}

\keywords{Deep Learning, Convergence, Data Augmentation, Adversarial Attacks Robustness}

\end{abstract}

\nomenclature[z-ANN]{ANN}{Artificial Neural Network}
\nomenclature[z-ML]{ML}{Machine Learning}
\nomenclature[z-DL]{DL}{Deep Learning}
\nomenclature[z-CV]{CV}{Computer Vision}

%% file: Contributions/contributions.tex
\begin{contributions}
In terms of original contributions, the thesis contains multiple instances of my ideas and work. The following techniques are based on my ideas, their implementation, experimentation, and analysis also consists of personal work:
\begin{enumerate}
    \item Perfect Ordering Approximation; a generalized model agnostic curriculum learning approach. The results show the effectiveness of the technique for improving training time as well as offer some new insight into the training process of deep networks.
    \item Cascading Sum Augmentation; an extension of mixup capable of utilizing more data points for linear interpolation by leveraging a smoother optimization landscape, a test time data augmentation technique is derived from this, improving adversarial attack robustness. This can be used for computer vision tasks in order to improve both prediction performance as well as improve passive model robustness.
\end{enumerate}

In order to show the improvements these approaches bring to the field of Deep Learning, I made use of the well known and generally accepted Computer Vision benchmark for image classification CIFAR.

The experiments compare the performance of models in similar conditions, with or without the usage of the techniques, and where applicable, are compared with results obtained by similar approaches.

I am also responsible for the implementation of the experimentation pipeline, which includes code for running and analyzing multiple experiments and making use of specialized hardware for half-precision processing.
I have also implemented various auxiliary tools used for data preprocessing and result visualization.

\end{contributions}

%% file: Chapter1/chapter1.tex



\nomenclature[z-AI]{AI}{Artificial Intelligence}
\nomenclature[z-AGI]{AGI}{Artificial General Intelligence}
\nomenclature[z-SOTA]{SOTA}{State-of-the-Art}
\nomenclature[z-NLP]{NLP}{Natural Language Processing}
\nomenclature[z-LM]{LM}{Language Model}

\chapter{Introduction \& Motivation}  

\ifpdf
    \graphicspath{{Chapter1/Figs/Raster/}{Chapter1/Figs/PDF/}{Chapter1/Figs/}}
\else
    \graphicspath{{Chapter1/Figs/Vector/}{Chapter1/Figs/}}
\fi

With an evergrowing presence of technology in our day to day life, machines are starting to have more impact on our lives. Humanity is delegating more responsibilities and decision makings to computers. This transition of influence to computers is moving at a fast pace while also being very hard for the general public to realize. At small obvious levels, we find smart wearables tracking the users' health in order to suggest exercise routines or warn about medical issues to smart calendars, synchronizing large companies or communities which send invitations in your name and help guide you to event destinations at the right time. Other approaches are more subtle, such as suggestion algorithms, which use your profile to offer people content they are likely to engage with, usually in the interest of the company. Moving over to macro levels, automated stock markets and Forex traders influence market conditions at a global scale, or weather prediction algorithms altering air or naval ship schedules. The trust and power we offer these systems are often paid back with great benefits to their users, increase security, financial gains, cost reductions and quality of life improvements are what humanity gets in return. Sadly, as it often is the case with powerful technology, mistakes and accidents can cause great damage, this is especially true when considering that bad actors will always find ways to use technology maliciously \cite{chesney2019deep}. For example, with the advent of social media, fake news began to be employed by ill-willing attackers. Considering this context we must become more aware of the shift of power from humans to machines so that we do not enter an exchange we will regret. Another thing to consider is that global trade, free travel, and increased access to technology has led to the creation and collection of an immense amount of data, further increasing the reach and applicability of the technology in our lives.

The current state of affairs in the domain of \textbf{Deep Learning} (DL) is dominated by the development and improvement of architectures and techniques designed to further push State-of-the-Art (SOTA) prediction performances. This direction was especially prominent in fields with direct real-world business applications such as Computer Vision (CV) and Natural Language Processing (NLP). 

One such recent notable example would be OpenAI's immense GPT-3 model \cite{brown2020language} which in its biggest form takes up to 350GB of memory for its 175 billion parameters. In order to train this deep network, industry specialists estimate the cost to be anywhere from \$4.000.000\cite{lambdaGPT3} and \$12.000.000\cite{tweet12MGPT3} which represents a 60 to 200 fold increase in cost compared to the previous version GPT-2. In the paper presenting this work, the researchers extend upon previous results showing that the power-law relation between compute power and validation loss of \textbf{Language Models} (LM)\cite{arisoy2012deep} holds true to even these orders of magnitude. Since the power-law relationship does not appear to be slowing down, we can rely on silicone manufacturing process shrinkage coupled with microprocessor architecture improvements in the coming years to push SOTA results by simply enabling the training of bigger models.

These kinds of advancements have even lead to multiple instances where deep learning approaches manage to beat humans in tasks that were traditionally considered too hard for AI to even compete. Examples of this can be seen in multiple fields such as natural language understanding \cite{wang2018glue} where the human baseline score was beaten as early as June 2019 \cite{liu2019multi}. Other impressive results can be seen in video games, where various research teams managed to train AI capable of defeating human world champions \cite{berner2019dota}, \cite{arulkumaran2019alphastar} in a task previously thought to be too hard for computers. Games like "DOTA 2" and "StarCraft II" present a serious challenge, since these are games with imperfect information and where long term planning is required. Passing this milestone and surpassing "simpler", perfect information games like Chess and Go brings us even closer to breakthroughs in tasks that present similar hurdles, such as autonomous driving.

While deep learning models are becoming bigger, more complex, and more powerful, because of their nested non-linear structure, these methods are usually developed and used only in a black-box manner. There is very limited understanding and we lack a theoretical framework capable of explaining what exactly makes deep networks arrive at their predictions \cite{8400040}. This lack of transparency is a major drawback which makes the use of potentially super-human AI unethical and potentially dangerous \cite{baum2018machine}, \cite{geis2019ethics}, \cite{arrieta2020explainable}, \cite{wachter2017transparent}, \cite{speicher2018unified} in fields such as medicine, governance, finance, policy, autonomous driving.

The research presented in this thesis is aimed to contribute to our understanding of deep learning models, as well as improving upon similar approaches. The focus of the two techniques is on convergence speed, data augmentation and adversarial attack robustness.

%% file: Chapter2/chapter2.tex
\chapter{Perfect Ordering Approximation}


\ifpdf
    \graphicspath{{Chapter2/Figs/Raster/}{Chapter2/Figs/PDF/}{Chapter2/Figs/}}
\else
    \graphicspath{{Chapter2/Figs/Vector/}{Chapter2/Figs/}}
\fi

\nomenclature[z-MSE]{MSE}{Mean Square Error}
\nomenclature[z-MAE]{MAE}{Mean Absolute Error}

\nomenclature[z-CL]{CL}{Curriculum Learning}

\section{Introduction \& Related Work}
Human learning is done both instinctively and consciously. Either one of the two ways of learning, from learning to crawl and walk to lecturing deep learning papers, they are usually done in a complexity increasing manner. Just as babies don't start learning how to walk without first mastering crawling, adults need to first understand the domain's fundamental concepts before they can begin to study advanced topics.

Starting from this observation of how learning seems to begin from simple concepts in order to grasp complex ones, the deep learning research community has proposed to replicate this procedure in training algorithms. The motivation for such a technique comes from the fact that gradually increasing difficulty seems to work very well for humans, in comparison with starting learning from a higher difficulty curriculum. If the same meta-learning behavior would hold for deep learning models, it should lead to both faster convergence speed as well as to higher final generalization performances.

The machine learning community has chosen to model this smart ordering of learning by replicating the human educational paradigm of students, teachers, and curriculum. \textbf{Curriculum Learning} (CL)\cite{Bengio_2009} is the name of the field of research which attempts to impose a certain structure on the training set itself. 

In the educational system, complex tasks are taught by teachers according to a curriculum. The curriculum itself is a guide on how to structure knowledge or tasks in an order which would result in the total information being assimilated successfully and as fast as possible. Usually, this ordering of learning is based on the complexity of the tasks. The students are gradually being introduced to more and more complex concepts and ideas. The teachers in these scenarios are entities that possess the information which is desired to be transferred and which can assess how difficult a certain task would be for the student at a given moment. Another observed behavior in human teaching is that we do not only sort learning items by complexity, but by how relatable and obvious some concepts seem to the student, we start from typical examples first and advance to more abstract and ambiguous ones \cite{avrahami1997teaching}.

In the framework of CL, the teacher is usually a pre-trained model which was trained without curriculum learning. This teacher model is responsible for either weighting the training samples based on their loss and ordering them based on this or by directly transferring the knowledge by the use of mechanisms such as soft-labels \cite{hacohen2019power}, \cite{dong2017multi}, \cite{weinshall2018theory}. The student in this framework being the model we are trying to train using CL.

While the CL framework and its many implementations have empirically proven to yield better results compared to the baseline and some recent limited theoretical results \cite{weinshall2018theory}, all of these start from the unproven assumption that the optimal way of ordering learning items is by difficulty. This assumption takes for granted that if we had access to an ideal difficulty scoring function (itself depends on an optimal hypothesis) for input data, we would achieve an optimal training procedure by gradually increasing the difficulty of data fed to the student. There are already conflicting empirical results that show that sometimes giving higher weights to more complex data-points leads to faster learning, as opposed to doing that for simpler samples. While the CL framework may discard other strategies that would outclass the increasing by difficulty strategies, it is both empirically and theoretically proven to beat the baseline random sampling training procedure. However, while a well-implemented CL should always beat the baseline, current techniques found in the literature depend on the existence of a pre-trained model with the role of the teacher, making the curriculum ordering/weighting decisions. This is a major limiting factor that makes CL practically applicable to a small subsample of tasks, such as multi-task learning, because having to train both a teacher and student is sometimes too expensive and yields similar results to simply training a single model for a longer period of time.

We believe that while the CL framework is able to produce impressive useful results \cite{graves2017automated}, \cite{weinshall2018curriculum}, \cite{Bengio_2009}, it is fundamentally limited from reaching or getting closer to an optimal training ordering strategy as long as the underlying assumption about the optimality of difficulty increasing strategies is unproven. Another limiting factor from the modeling of this framework is the requirement of defining teachers, a computationally prohibitive characteristic.

To state it clearly, it may be the case that the underlying assumptions of optimality prove to be true in which case we won't require any other framework, except the CL framework for this problem. It might also be the case that any other strategy resulting from different frameworks are similar enough to the best CL approaches that it might not be worth using, or that others, closer to optimal strategies will prove to be significantly more prohibitive in terms of compute or memory requirements, and will only remain in the reals of academic experimentation with no real-world applicability.

In this work, we will attempt to approach the problem of data structuring without depending on the same limiting factors as CL or the same assumption.

\section{Perfect Ordering}
This work relies on the following unproven assumption: "Given an artificial neural network and a dataset there must be at least one optimal order of providing training items to the model". While we are not capable of proving this assumption in a theoretical framework, other works on CL \cite{weinshall2018theory} and empirical results provide enough circumstantial evidence to warrant the exploration of the problem from this standpoint.

To motivate the usefulness of doing this work, we must hold another assumption: "Having an optimal ordering of training items for a given neural network and dataset, the predictive power of the resulting trained network must be significantly higher than training the model in a random ordering fashion or that the optimal ordering reaches the same convergence plateau in a smaller number of gradient steps". We will be providing experimental results that indicate that the above assumption is true, at least in some scenarios.

We will be looking at optimality in terms of the speed of convergence to an accuracy plateau, meaning that we are trying to find the shortest series of batches we have to train on that reaches convergence.

To this end, my colleague Cosmin Pascaru has devised the following experiments in order to empirically validate our Perfect Ordering assumptions. Using a subset of the MNIST dataset of only 24 images of zeroes and 24 images of ones for training, and 2048 total testing samples, using batches of size 8, we train all possible batch permutations. All runs used the same initial weight initialization of a small convolutional network and the same batch split for each epoch. We do this in order to ensure that all conditions are identical.

The dataset used had to be small enough in order for the experiments to be computationally feasible, to that end, we will only be considering all possible orderings of batches for a fixed batch split. Since the number of all such possible orders \textbf{for a single epoch} is:

\[ NrOrd_{batches}(n, b) = (n/b)! \]

Where n is the number of samples and b is the batch size. Resulting from the above formula, the experiment had to test 720 permutations of batches for the first epoch.

In order to reduce the immense number of potential permutations when multiple epochs are concerned, Cosmin utilized the KMeans algorithm to obtain 12 runs from the previous epoch, by clustering the results by their test loss. This leads to only $ 720 + 12 \cdot 720 \cdot 12 = 104400 $ iterations to use for 13 epochs. The algorithm proposed by him can be seen below:

\begin{algorithm}[H]
    $InitNet \gets$ Initialize the network with random weights\;\\
    $Batches[0] \gets$ Generate a random split into batches\;\\
    \For{$Perm$ \KwIn $permutations(Batches[0])$}{
        $Nets[0].append(train(InitNet, Perm))$\;
    }
    \For{$epoch\gets1$ \KwTo $NrEpochs-1$}{
        $Batches[epoch] \gets$ Generate a random split into batches\;\\
        $StartNets \gets {KMeansSelectByTestLoss}(Nets[epoch - 1],\ clusters=12)$\;\\
        
        \For{$Perm$ \KwIn $permutations(Batches[epoch])$}{
            \For{$StartNet$ \KwIn $StartNets$}{
                $Nets[epoch].append(train(StartNet, Perm))$
            }
        }
    }
    \caption{Train with all batches combinations \label{alg:all_batches_comb}}
\end{algorithm}

\begin{figure}[H]
\centering
\begin{subfigure}{.5\textwidth}
  \centering
  \includegraphics[width=1\textwidth]{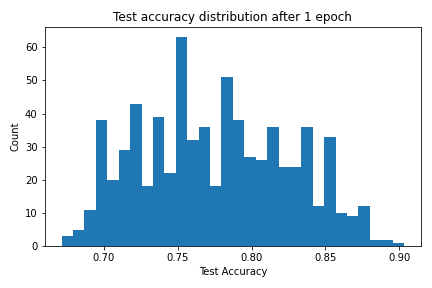}
\end{subfigure}%
\begin{subfigure}{.5\textwidth}
    \centering
    \includegraphics[width=1\linewidth]{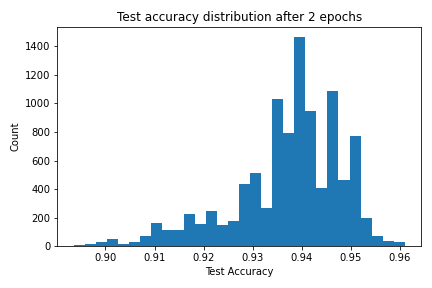}
\end{subfigure}\\%
\begin{subfigure}{.5\textwidth}
    \centering
    \includegraphics[width=1\linewidth]{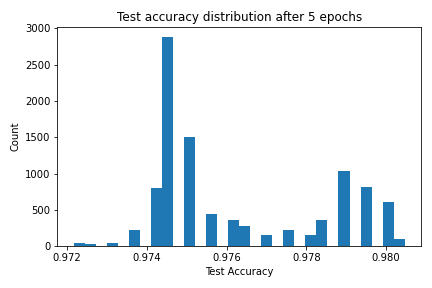}
\end{subfigure}%
\begin{subfigure}{.5\textwidth}
    \centering
    \includegraphics[width=1\linewidth]{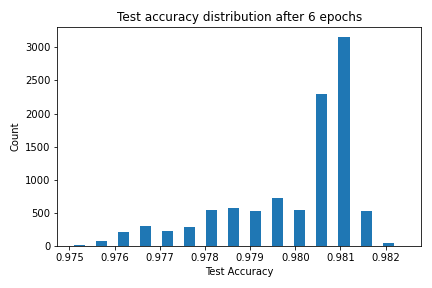}
\end{subfigure}%
\caption{Distribution of test accuracy after 1, 2, 5, and 6 epochs\label{fig:dist_losses_1256}}
\end{figure}

Figure \ref{fig:dist_losses_1256} shows the distribution of test losses after 1, 2, 5 and 6 epochs, respectively. As Cosmin states in his Master's Thesis, the plots seem to follow a Gaussian distribution, which gets is biased towards the higher values.

\begin{figure}[H]
\centering
\begin{subfigure}{.5\textwidth}
  \centering
  \includegraphics[width=1\linewidth]{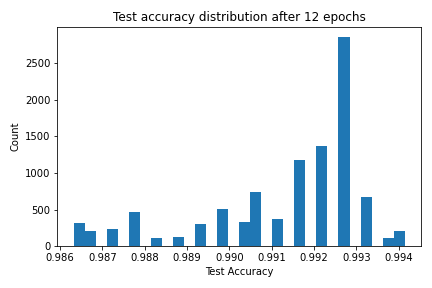}
\end{subfigure}%
\begin{subfigure}{.5\textwidth}
    \centering
    \includegraphics[width=1\linewidth]{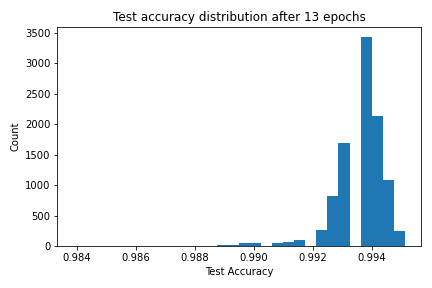}
\end{subfigure}%
\caption{Distribution of test accuracy after 12 and 13 epochs\label{fig:dist_losses_12_13}}
\end{figure}

Figure \ref{fig:dist_losses_12_13} shows the test accuracy distributions for the final epochs, 12 and 13. These are very similar to the ones earlier in the training, but by looking at the range of values, they are much tighter.

Another observation is that with more epochs, the interval of possible values shrinks, this is expected when taking into consideration the lottery ticket hypothesis \cite{frankle2018lottery} since when the models begin to pick a "ticket" and begin to utilize only a small network subset, the potential variations in how data is processed in the network diminishes.

An important result from this analysis is that the maximum accuracy after $N$ epochs is greater than the minimum accuracy after $N+4$ epochs, with $MaxAcc(4) = MinAcc(10)$ in the most extreme case. This is important since when relying on completely random sampling to generate our batches and order, it is just as likely to obtain a configuration that leads to the worse performance possible, as well as the best one.

The maximum and minimum accuracy and loss can be seen bellow in table \ref{tab:summary} for all 13 epochs.

It is interesting to notice that in virtually all cases, the maximum accuracy after $N$ epochs is higher than the mean accuracy after $N+1$ epochs. 

\begin{table}[h]
\centering
\begin{tabular}{l c c c c c c}
\cmidrule[2pt]{1-6}
\multirow{1}{*}{\textbf{Epoch}} & \multicolumn{1}{c}{Min Accuracy} & \multicolumn{1}{c}{Mean Accuracy} & \multicolumn{1}{c}{Max Accuracy} & \multicolumn{1}{c}{Min Loss} & \multicolumn{1}{c}{Max Loss} \\
\cmidrule{1-6}
	\hfill 1 & 0.671 & 0.776 & 0.903 & 2.393 & 2.449 \\
	\hfill 2 & 0.894 & 0.937 & 0.961 & 1.643 & 1.714 \\
	\hfill 3 & 0.959 & 0.966 & 0.971 & 0.983 & 1.023 \\
	\hfill 4 & 0.965 & 0.973 & 0.976 & 0.626 & 0.767 \\
	\hfill 5 & 0.972 & 0.976 & 0.980 & 0.431 & 0.582 \\
	\hfill 6 & 0.975 & 0.980 & 0.982 & 0.322 & 0.429 \\
	\hfill 7 & 0.979 & 0.982 & 0.985 & 0.254 & 0.294 \\
	\hfill 8 & 0.976 & 0.983 & 0.987 & 0.217 & 0.324 \\
	\hfill 9 & 0.979 & 0.987 & 0.990 & 0.190 & 0.294 \\
	\hfill 10 & 0.976 & 0.989 & 0.992 & 0.178 & 0.345 \\
	\hfill 11 & 0.979 & 0.990 & 0.993 & 0.162 & 0.320 \\
	\hfill 12 & 0.986 & 0.991 & 0.994 & 0.144 & 0.237 \\
	\hfill 13 & 0.984 & 0.994 & 0.995 & 0.126 & 0.269 \\
\cmidrule{1-6}
\end{tabular}
\caption{Summary of results on test data after each epoch\label{tab:summary}}
\end{table}

The results from Cosmin Pascaru's experiment and analysis indicate that our previous assumptions seem to hold true. We want to accentuate the fact that these results were obtained on a very small dataset, training a toy convolutional neural network. This model converges to an accuracy plateau after 12 epochs, and each epoch consists of a very small number of gradient steps.

\section{Perfect Ordering Approximation}
Having had the previous analysis, we have gained some insight into just how important a factor learning ordering is. Similarly, with weight initialization, which wasn't given much thought during the early stages of ANN history, we consider that learning ordering is a facet of the training procedure, which is currently causing us to miss out on important convergence speed gains. The vast majority of ANN approaches today will employ smart parameter initialization schemes, but will rely on random uniform batch sampling from the dataset. This could be due to multiple causes, some we believe to be the main ones, is the fact that there isn't a very clear understanding in the DL field of how much a smarter learning ordering strategy would beat the random approach. Another factor is that all current strategies come with a considerable computational overhead which proves to cost too much for them to be viable from a cost-benefit perspective for the majority of tasks, a clear exception being multi-task challenges for which CL becomes computationally viable. 

We want to breach the gap of understanding of how important smart ordering strategies can be and to propose some techniques which could lead to becoming computationally viable to use for most ML tasks. Ideally, we would reach a point in the future where all ANN approaches will employ a smart ordering strategy paying an insignificant cost compared to the upside in convergence speed, this would make them be used as a default go-to tool, just all works utilize weight initializations, which are better thought out and understood than uniform sampling. Realistically, we are currently aiming to find a technique that, while computationally expensive, offers enough speed of convergence in the very start of training (where the effects of better ordering are the greatest) to be viability used as a "warm-up" method. As a side note, weight initialization has reached a point of normality where many research papers will even forget to mention which technique they used, this can be found in quality papers as well.

In this work, we will present the following strategies we have attempted in order to approximate the optimal \textbf{"Perfect Ordering"}, which was achieved in a brute-force manner. In order to conceptualize the framework we have developed and ran these experiments, we will use the following terms and concepts which compose our \textbf{Perfect Ordering Approximation} framework:
\begin{itemize}
    \item \textbf{Learning Item:} this will refer to either an individual data sample or to a mini-batch of samples. We do this unification of terms because we want our other mechanism to be more generalized and to be able to work on either individual samples or batches of data;
    \item \textbf{Scoring Mechanism:} this is a function that will generate a score or a label for a given learning item. This score could be generated with or without the use of the model we are training, it can be a simple metric such as a loss value, or a label;
    \item \textbf{Sampler Strategy:} this is an algorithm which given the scores or labels obtained using a scoring mechanism will construct a batch based on a strategy. Some simple sampler strategies are choosing the smallest scores, choosing the largest scores, sampling uniformly based on the scores. By their nature, sampler strategies construct batches from the dataset with replacement, meaning that the training procedure itself would not use epochs anymore since the sampler will simply generate batches when asked to, the same learning items can be chosen multiple times consecutively and some might never be used at all;
    \item \textbf{Ordering Strategy:} this is an algorithm that will sort a given list of scores/labels based on a given strategy and will output the first learning item associated with the first score in the sorted list. The most obvious ordering strategies would be sorting the scores in an ascending or descending manner, and will then output the first learning item, on the second call of the function the next learning item will be returned; 
    \item \textbf{Data Loader:} this is the central control mechanism in our framework. A data loader has a scoring mechanism, and either a sampler or an ordering strategy associated with it. The data loader is the one responsible for choosing when to update the scores of learning items using the scoring mechanism, for example, we might only update scores once per epoch, or we could update scores after each gradient update. The data loader is the one which feeds the list of scores and associated data to the sampler/ordering strategy algorithm and then uses the result in the main training loop to train the model with that learning item;
\end{itemize}

\section{Experiments}
The experiments we designed, follow the general principle of replicating the same conditions as the most generic architecture, dataset, and meta-parameters usually encountered in literature. We value this because ideally, a working \textbf{Perfect Ordering Approximation} strategy should apply to any given configuration. So even if other datasets or meta-parameter values would probably have yielded better results for some strategies, we chose to utilize the same configuration across all experiments:
\begin{itemize}
    \item The CIFAR-10 dataset;
    \item  Pre-Activation ResNet with Identity Mapping \cite{he2016identity} with 56 layers (PreResNet-56);
    \item SGD optimizer with a learning rate of 0.05, L2-normalization factor of 0.0001, Nesterov \cite{sutskever2013importance} momentum of 0.9;
    \item A mini-batch size of 128;
    \item A pre-processing pipeline where we normalize the data and augment it with random flips and crops;
    \item The learning rate was halved every 30 epochs;
    \item The loss function is the cross-entropy loss, which in PyTorch will apply softmax on the output internally;
\end{itemize}

The experiments were set up to use the same seed for randomization, as such, the model parameter initialization, an initial data shuffle, was identical across runs for a better comparison.

We will be describing the \textbf{Perfect Ordering Approximation} in the context of the concepts and framework presented previously.

\subsection{Sample Loss Orderer \& Sampler}
Perhaps the simplest approach one could attempt in order to structure the training data. The intuition being that the loss of a sample contains information about how the model itself perceives said sample, an indication of difficulty the data presents to the model.

This strategy has the following framework composition:
\begin{enumerate}
    \item \textbf{Learning Item}: an item for the sampling strategy variant, a batch for the ordering strategy variant;
    \item \textbf{Scoring Mechanism}: forward pass the learning item through the network and calculate the loss, this is the score assigned to the item;
    \item \textbf{Sampling Strategy}: randomly extract a batch of samples, the probability an item is chosen is weighted by its score. Both direct weighing and inverse weighing were considered;
    \item \textbf{Ordering Strategy}: sort the indices of the learning items using their scores, and return the first batch of samples in the list, eliminate the samples from the list for the next function calls. Both ascending and descending orderings were considered;
    \item \textbf{Data Loader}: In order to prove how efficient or inefficient this strategy is, the data loader recalculates the scores after each gradient step update. In order to make this strategy computationally viable, the data loader randomly selects $K$ learning items to consider. The algorithm only scores these batches, and the strategies will only order/sample from these items;
\end{enumerate}

In our experiments, using this scoring mechanism, both the sampling and ordering strategies lead to very similar results, and due to this, we will only be showing plots of ordering strategy results as they are indistinguishably similar to the sampling equivalent.

In figure \ref{fig:multi_attempt_loss} you can see the train and test accuracies for the loss scoring, ascending ordering strategy runs:

\begin{figure}[H]
\centering
\begin{subfigure}{.5\textwidth}
  \centering
  \includegraphics[width=1\linewidth]{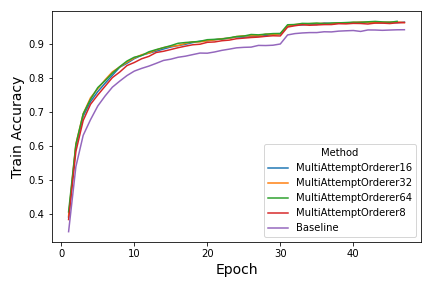}
\end{subfigure}%
\begin{subfigure}{.5\textwidth}
    \centering
    \includegraphics[width=1\linewidth]{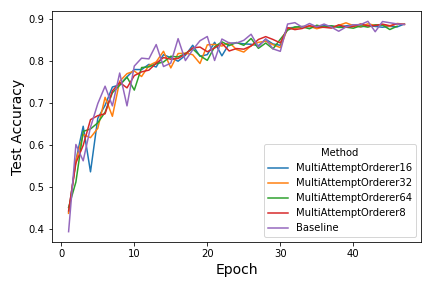}
\end{subfigure}%
\caption{Multi Attempt Loss Orderer Results \label{fig:multi_attempt_loss}}
\end{figure}
This \textbf{Multi Attempt Loss Orderer} strategy has been executed with multiple values for  $K=\{8,16,32,64\}$. As can be seen, all the runs that used the strategy produce better training accuracies, but this apparent boost in performance does not translate in on the test data. While not producing a clearly better model, it does seem that the baseline result experienced more variance than the ordering strategy.

When taking a closer look at the first epochs, which are the most important in evaluating these methods, we do see a noticeable difference for the first epoch between the baseline result and the other, but it seems that any benefit from increasing $K$ becomes insignificant. A few of the starting epochs can be seen in table \ref{tab:multiattemptloss}.

\begin{table}[h]
\centering
\begin{tabular}{l c c c c c c}
\cmidrule[2pt]{1-6}
\multirow{1}{*}{\textbf{Epoch}} & \multicolumn{1}{c}{Baseline} & \multicolumn{1}{c}{MA K=8} & \multicolumn{1}{c}{MA K=16} & \multicolumn{1}{c}{MA K=32} & \multicolumn{1}{c}{NA K=64} \\
\cmidrule{1-6}
    \hfill 1  & 0.3926 & 0.4422 & 0.4392 & 0.4363 & 0.4503  \\
    \hfill 2  & 0.6005 & 0.5564 & 0.5622 & 0.5613 & 0.5109  \\
    \hfill 3  & 0.5616 & 0.5984 & 0.644  & 0.624  & 0.6318  \\
    \hfill 4  & 0.6415 & 0.6598 & 0.5353 & 0.6168 & 0.6375  \\
    \hfill 5  & 0.698  & 0.6693 & 0.6631 & 0.6391 & 0.6522  \\
    \hfill 6  & 0.7395 & 0.6737 & 0.6938 & 0.713  & 0.6775  \\
    \hfill 7  & 0.6926 & 0.7283 & 0.7376 & 0.6677 & 0.7238  \\
    \hfill 8  & 0.7709 & 0.7485 & 0.7426 & 0.7512 & 0.7417  \\
    \hfill 9  & 0.6926 & 0.7355 & 0.7618 & 0.7694 & 0.7611  \\
    \hfill 10 & 0.7875 & 0.7648 & 0.7792 & 0.7757 & 0.7302  \\
\cmidrule{1-6}
\end{tabular}
\caption{Accuracy results on test data for the baseline and the Multi Attempt approaches \label{tab:multiattemptloss}}
\end{table}
As can be seen, there is no noticeable difference between the baseline and the other runs except the fact that the baseline accuracy values seem to alternate more.

This result is expected since we are using the insight a currently untrained model has regarding what it considers to be the "easiest" samples to train on. Curriculum Learning has to face the same issue, they resolve it by using a pre-trained model as a teacher, the teacher model having a more accurate prediction of the complexity of a given data point. The fact that the values for the baseline are much more volatile than the Multi Attempt method is also explainable by the fact that by restricting the possible number of ordering configurations, the model is resulting in model performance, and is also more constrained. Another cause for this behavior might be the fact that the losses propagated in the model are small, hence, the gradients will be small, as well leading to smaller gradient steps, the model itself not moving as much in the weight space, it is interesting that it is able to maintain roughly similar performance even with this limitation.

What was unexpected is the fact that this method seems to encourage the memorization of data since it increased the training accuracy while not affect testing.

We have also experimented with a decreasing ordering of the losses. In this scenario, the model was incapable of learning anything, no matter how many iterations passed. We speculate that this is because the model makes big gradient steps due to the very large loss value and it keeps bouncing between extreme parameter weight values. 

\subsection{Maximum Loss Delta Orderer}
The second strategy we tested takes inspiration from how humans might evaluate the complexity of a learning task. We want to evaluate the "usefulness" of learning a certain item by how much we were able to learn from it, which we believe to be a much better rule of thumb to go by than simply looking at increasing complexity over time. For example, while it might be easy to learn to do a simple magic trick, a CL approach would perhaps choose this as one of the first tasks to be processed, but learning a magic trick might simply not be of interest to our goal of being a better tennis player. On the opposite end, a task might prove to be very difficult, a CL approach would not choose to teach us this item but it might be the case that while difficult, the benefit from learning that concept is big enough to make it worthwhile.

Starting from this human intuitive concept, we designed a Perfect Ordering Approximation strategy that would replicate this concept. Below you can find the framework concepts explanation for the \textbf{Maximum Loss Delta Orderer}:
\begin{enumerate}
    \item \textbf{Learning Item}: a batch of samples;
    \item \textbf{Scoring Mechanism}: take the learning item and do a gradient update step, using it and store the loss value associated with that batch or with another static pre-chosen batch. Do another forward pass in the network with the previous batch and calculate the difference between the previous loss and the new one, this value is the score associated with the learning item. After each scoring, the model weights and SGD momentum parameters need to be rolled back to their previous state;
    \item \textbf{Ordering Strategy}: sort the indices of the learning items using their scores, and return the first batch of samples in the list, eliminate the samples from the list for the next function calls. Only a descending order is considered;
    \item \textbf{Data Loader}: In order to prove how efficient or inefficient this strategy is, the data loader recalculates the scores after each gradient step update. In order to make this strategy computationally viable, the data loader randomly selects $K$ learning items to consider. The algorithm only scores these batches and the strategies will effectively return the maximum loss difference before and after updating the parameters;
\end{enumerate}
The intuitive idea is that if by updating the parameters of the model using a batch, we reach a better weight state, we measure just how much taking this gradient step would improve the model's loss.

Since we are selecting new $K$ batches to score after every gradient step, we are effectively selecting the maximum loss drop, which would be obtained by teaching the model those $K$ items.

For this, we tested two ways of evaluating calculating the difference in loss. The first one is by using the same sample we are making an update with (\textbf{Same Batch Loss Delta Variant}). Meaning that we would first store the loss obtained by the learning item, update the model using it, and then forward pass the same batch again to compare the new loss value to the old one. 

The second method was by using a static batch pre-sampled before we start scoring the learning items. We chose to randomly sample a batch of 512 images from the test dataset at the start of each epoch (\textbf{Static External Batch Loss Delta Variant}). This way, we aim to directly look at how much the appliance of a certain gradient step would help our model in its new date prediction performance.

\subsubsection{Same Batch Loss Delta Variant}

\begin{figure}[H]
\centering
\begin{subfigure}{.5\textwidth}
  \centering
  \includegraphics[width=1\linewidth]{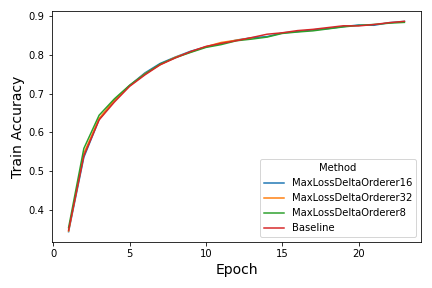}
\end{subfigure}%
\begin{subfigure}{.5\textwidth}
    \centering
    \includegraphics[width=1\linewidth]{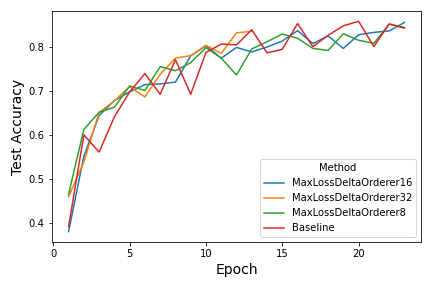}
\end{subfigure}%
\caption{Maximum Loss Delta Orderer Results \label{fig:max_delta_loss}}
\end{figure}

In figure \ref{fig:max_delta_loss}, the train and test accuracies of the method use the learning item's previous and new loss delta as scores. You can see the accuracy results for runs with values for $K$ of 8, 16, or 32.

Using this data structuring strategy, we obtain some interesting results. The first one being that the configurations seem to behave identically with the baseline on the trained dataset. However, they perform quite differently on the test data. Again, we see that the baseline run is more sporadic and can fluctuate with periods of continuous decreases in accuracy followed by big spikes in performance. The usage of \textbf{Maximum Loss Delta Orderer} seems to induce the same behavior of constraining the model from varying too much in its evolution across training. With this technique also, we see that all runs will hover close values in accuracy the longer we train it.

Due to computational budget constraints, we had to limit the number of epochs. We trained our model based on the value of $K$, with our current proof-of-concept implementation, the running time of our algorithm scales linearly with $K$. Because of this, the run with $K=32$ was only left to run for 12 epochs, the run with $K=16$ for 24 epochs and the run with $K=8$ made use of 48 training epochs.

With this technique, the gains obtained from increasing $K$ are even smaller than the previous strategy, as such, we foresee an efficient implementation of this algorithm with $K=8$ to be able to use up just as much time by using more memory and GPU compute units.

Just as the previous strategy, we see its applicability coming in for the first epochs or gradient steps of training, where it can be used as a warm-up algorithm to more quickly get the model at a more stable state for any given training configuration. While there might still be some applicability due to the apparent increase in training stability, which could be valuable for certain scenarios, The test accuracy for the first 10 epochs of training can be seen in table \ref{tab:maxdeltaloss}.

\begin{table}[h]
\centering
\begin{tabular}{l c c c c c}
\cmidrule[2pt]{1-5}
\multirow{1}{*}{\textbf{Epoch}} & \multicolumn{1}{c}{Baseline} & \multicolumn{1}{c}{MLD K=8} & \multicolumn{1}{c}{MLD K=16} & \multicolumn{1}{c}{MLD K=32} \\
\cmidrule{1-5}
    \hfill 1  & 0.3926 & \textbf{0.4663} & 0.4211 & 0.4603  \\
    \hfill 2  & 0.6005 & \textbf{0.6129} & 0.5521 & 0.5388  \\
    \hfill 3  & 0.5616 & \textbf{0.6518} & 0.6443 & 0.6515  \\
    \hfill 4  & 0.6415 & \textbf{0.6633} & 0.6785 & 0.6776  \\
    \hfill 5  & 0.698  & \textbf{0.7114} & 0.6981 & 0.7101  \\
    \hfill 6  & \textbf{0.7395} & 0.7011 & 0.7144 & 0.6866  \\
    \hfill 7  & 0.6926 & \textbf{0.7552} & 0.7162 & 0.7379  \\
    \hfill 8  & 0.7709 & 0.7461 & 0.7202 & \textbf{0.7747}  \\
    \hfill 9  & 0.6926 & 0.7643 & 0.7797 & \textbf{0.7802} \\
    \hfill 10 & 0.7875 & 0.7983 & 0.8025 & \textbf{0.8037}  \\
\cmidrule{1-5}
\end{tabular}
\caption{Accuracy results on test data for the baseline and the Max Loss Delta Orderer \label{tab:maxdeltaloss}}
\end{table}

As can more clearly be seen in the \ref{tab:maxdeltaloss} table, the fluctuations of accuracy are far greater with the baseline run compared to the MLD method. While the baseline has obtained a better accuracy score on epoch 6, it seemed to be a spike which value doesn't consistently reach, while our method does still fluctuate, it is by a far more reduced margin.

Our method seems to make the most sense when applied for the first epochs for an overall better starting point and to be used through the training when model stability is a major consideration.

Another variation of the strategy we ran was to score the learning items by the relative drop in loss, not the absolute. The formula for that would be:
\begin{equation*}
    score = \frac{ prevLoss - newLoss }{prevLoss}
\end{equation*}

This variation leads to a slightly worse performance compared to the absolute difference version but would still beat the baseline in the first epochs and stability.

In order to see what an upper limit of this strategy would be in regard to values of $K$, we also trained 1 epoch where $K=391$ (the number of batches you obtain from CIFAR-10 with a batch size of 128). This run obtained a test data accuracy score of \textbf{0.4715}, which is better than the smaller values for $K$, but not sufficient enough to warrant the 11 hours of training time required to finish this 1 epoch. Even with the envisioned parallelization optimizations, we only foresee a \textbf{x8 boost} in speed when using a card similar to the RTX 2080Ti.

\subsubsection{Static External Batch Loss Delta Variant}

As described previously, we also implemented a variant that will use a biggest batch of 512 images samples from the test set at each epoch start. We predicted that by guiding our ordering decision directly by the end goal of the task, we would obtain even greater convergence boosts.

\begin{figure}[H]
\centering
\begin{subfigure}{.5\textwidth}
  \centering
  \includegraphics[width=1\linewidth]{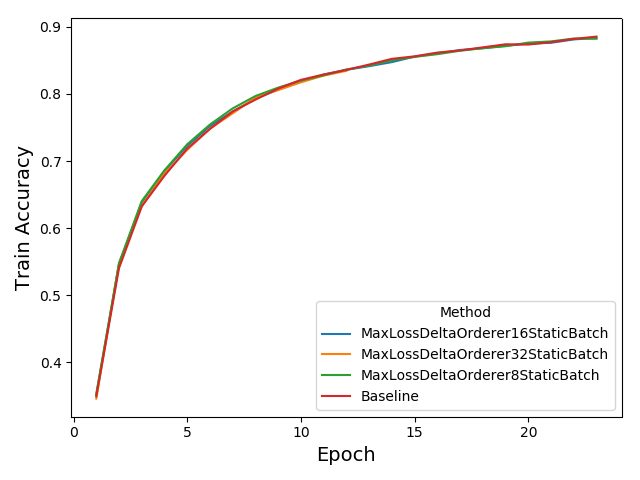}
\end{subfigure}%
\begin{subfigure}{.5\textwidth}
    \centering
    \includegraphics[width=1\linewidth]{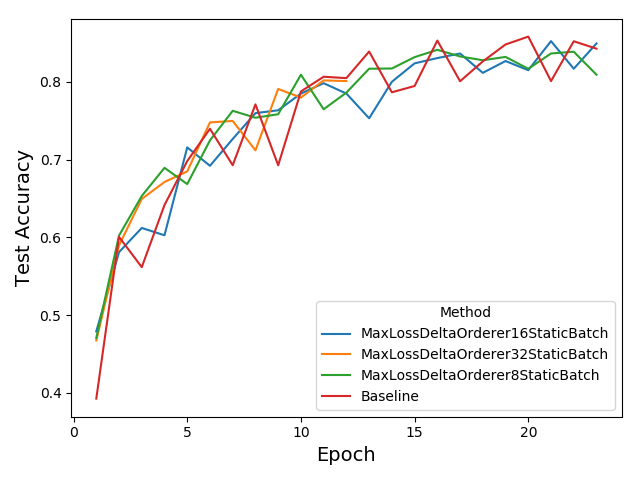}
\end{subfigure}%
\caption{Maximum Loss Delta Orderer Results \label{fig:max_delta_loss_test}}
\end{figure}

In figure \ref{fig:max_delta_loss_test}, the train and test accuracies of the method which uses the learning item's previous and new loss delta as scores. You can see the accuracy results for runs with values for $K$ of 8, 16, or 32.

We can see that they do produce very similar results, training being identical to the baseline and testing starting from a higher point in the starting epochs. The stability increase compared to the baseline seems to hold true here as well.

The test accuracy for the first 10 epochs of training can be seen in table \ref{tab:maxdeltalosstest}.

\begin{table}[h]
\centering
\begin{tabular}{l c c c c c}
\cmidrule[2pt]{1-5}
\multirow{1}{*}{\textbf{Epoch}} & \multicolumn{1}{c}{Baseline} & \multicolumn{1}{c}{MLD K=8} & \multicolumn{1}{c}{MLD K=16} & \multicolumn{1}{c}{MLD K=32} \\
\cmidrule{1-5}
    \hfill 1  & 0.3926 & 0.4705 & \textbf{0.479}  & 0.4673  \\
    \hfill 2  & 0.6005 & \textbf{0.6022} & 0.5809 & 0.5897  \\
    \hfill 3  & 0.5616 & \textbf{0.6531} & 0.612  & 0.6493  \\
    \hfill 4  & 0.6415 & \textbf{0.6893} & 0.6026 & 0.6711  \\
    \hfill 5  & 0.698  & 0.6684 & \textbf{0.7156} & 0.6847  \\
    \hfill 6  & 0.7395 & 0.7246 & 0.6919 & \textbf{0.7477}  \\
    \hfill 7  & 0.6926 & \textbf{0.7626} & 0.7264 & 0.7496  \\
    \hfill 8  & \textbf{0.7709} & 0.7537 & 0.7596 & 0.7118  \\
    \hfill 9  & 0.6926 & 0.7582 & 0.7633 & \textbf{0.7907}  \\
    \hfill 10 & 0.7875 & \textbf{0.809}  & 0.7848 & 0.8017  \\
\cmidrule{1-5}
\end{tabular}
\caption{Accuracy results on test data for the baseline and the Max Loss Delta Orderer \label{tab:maxdeltalosstest}}
\end{table}

Looking closely at the test accuracies over the first 10 epochs, we see that this variation outperforms the other one for all values of $K$ for the majority of epochs, the very first epoch being the most important one since it is the one least effected by random chance. While we still do not see a clear, major difference between the baseline and our strategy after the first epoch, the stability of having almost constant improvement over the last epoch and no random spikes and drops is even more pronounced now.

Since we could obtain an efficient parallelised algorithm which if given enough GPU memory and compute units should reach similar running time per epoch as the baseline for values of $K\leq 8$. Paired with the clear advantage on generalization in the first epochs, makes our strategy a good candidate to be used as a default warm-up technique which should work on any model or dataset since it simply looks at loss values.

 When we trained 1 epoch with $K=391$, we obtained only a slightly higher potential gain compared to the previous variation, this variation obtained a test data accuracy score of \textbf{0.4831}. While still better the predictive gain obtained compared to the $K=8$ run is too small to justify the several orders of magnitue longer training time.

\section{Conclusions}
This work started with the results obtained by Cosmin Pascaru regarding the importance of learning item ordering in training an artificial neural network. We presented why we think that the Curriculum Learning framework might be excluding much better data structuring strategies and we presented our \textbf{Perfect Ordering Approximation} framework. 

Using this framework, we implemented and experimented with several strategies, such as loss ordering and loss delta ordering. Some of which show great promise for usage as a warm-up technique and others, which did not yield good results but which offered us some more insight regarding the deep learning training procedure. Some important factors to be remembered are that the models trained with some of these techniques appear to have a more stable training evolution and that all of these techniques do not require to first train a teacher model, but can structure the data in a meaningful manner using the model we are trying to teach. An interesting future work would be to combine methods originating from the two frameworks CL being the one which leverages the teacher models knowledge, our methods leverage the student model.

The \textbf{Perfect Ordering} analysis was done on a very easy task, we should firstly try to design a slightly more complex task, dataset, and model configuration. This should come accompanied by a parallelization of the brute-force algorithm such that it is feasible to run the experiment in a few days on multiple machines with multiple cards. This should be the first step moving forward, having done that we should then attempt another analysis of the results, looking at as many metrics or relations as we could think of. The scope of this more detailed analysis on a more complex dataset should try and identify a pattern over certain metrics or a combination of metrics in the best runs compared to the worst ones or the median case. If such patterns are identified, we should be able to move ahead and implement an efficient approximation of those metrics which could be used as a warm-up for potentially any task or dataset. 

The current implementation is very much a proof-of-concept in the sense that no optimizations regarding computational requirements or parallelization have been applied. Multiple obvious optimizations can be applied, the most important one being parallelization of the scoring of learning items. This optimization would make the algorithm be able to speed up score calculations by about 8 times on a NVIDIA RTX 2080Ti (based on how many separate models we were able to train in parallel on the card). This speedup is a lower bound, considering we would simply clone the model, data loaders, optimizers, etc.

Another important future work would be to repeat the experiments on at-least 5 different seeds. While the methods used reduce randomness between runs by their very nature, in order to obtain further confidence in using these methods, we would need to acquire and deploy additional computational power. A factor that could introduce variance in the results come from the initialization, while unlikely it is possible that the initial parameter values lead themselves to be especially attuned to our methods. Another major influence could come from the randomness of the learning items picked for scoring, having multiple epochs reduced the effect a potentially lucky draw of samples could have on the overall performance, but additional runs would increase confidence in the results.

%% file: Chapter3/chapter3.tex
\chapter{Cascading Sum Augmentation}


\ifpdf
    \graphicspath{{Chapter3/Figs/Raster/}{Chapter3/Figs/PDF/}{Chapter3/Figs/}}
\else
    \graphicspath{{Chapter3/Figs/Vector/}{Chapter3/Figs/}}
\fi

\nomenclature[z-GCC]{GCC}{Gradually Cascading Coefficient}

\section{Introduction} \label{introduction}
Data augmentation techniques help deep artificial neural networks improve their generalization performance. Traditional image augmentation approaches are biased towards using methods that generate samples that are comprehensible for a human observer. This tendency of favoring understandable images leads to ignoring a class of data that a neural network could benefit from during the learning process. In this paper, we present a simple data augmentation technique for image classification tasks called \textbf{Sum Augmentation}, which builds upon previous work regarding linear interpolation of inputs. We are able to usefully combine a significantly larger number of data points, thus dramatically increasing the size of the dataset. We showcase how this method can be used to linearly combine up to 8 samples to produce seemingly useless images, from which a deep model is able to extract useful knowledge. We introduce a novel training procedure, called \textbf{Cascading Sum Augmentation}, which successfully transfers this knowledge to reach a greater generalization ability. We find that the accuracy gain is greater for smaller datasets. A novel test-time augmentation procedure is derived from \textbf{Sum Augmentation}, to further increase performance.

Recent advances in Deep Learning have greatly improved performance on various Computer Vision tasks, including classification, object detection, and image segmentation. However, training Deep Artificial Neural Networks requires a substantial amount of labeled data, which may prove costly or impractically hard to obtain, especially in certain domains such as robotics or medicine. This searches for ways to artificially augment the data we have in order to obtain more generalized models a worthwhile endeavor. One very important argument is that security is extremely relevant in today's world \cite{barreno2010security} \cite{barreno2006canmlsecure}, and neural networks with superior generalization power not only produce better results, but they are also more resistant to adversarial attacks.  Many data augmentation techniques have been proposed, each with various degrees of success and use cases, depending on the data sets, models, and machine learning techniques used. They are at the heart of many applications, ranging from image classification \cite{krizhevsky2017imagenet} to speech recognition \cite{graves2013speech} \cite{amodei2016deepspeech}. The majority of data augmentation algorithms used result in images that are understandable and can be correctly identified by a human observer e.g. horizontal/vertical flipping, cropping, rotation, brightness adjustments \cite{goyal2017accurate} \cite{lecun1998gradient} \cite{simonyan2014verydeep} or even random erasing \cite{zhong2017random}.

Recent work in the field of image data augmentation \cite{zhang2017mixup}, \cite{tokozume2018between}, \cite{tokozume2017learning}, \cite{inoue2018japaneseguy}) showed that techniques outside the class of augmentations which obtain images with humanly perceivable characteristics are not only learnable by deep neural networks, but they also improve the generalization performance of said networks.

In this paper we present a generalized method of generating linear combinations of data points called \textbf{Sum Augmentation} (section \ref{sum_augmentation}), which has proven to greatly reduce error rates on CIFAR-10/100. 

\section{Related Work} \label{related_work}
Chawla et al.\cite{chawla2002smote} propose the augmentation of an imbalanced dataset by interpolating between the nearest neighbors of the rare class. DeVries et al. \cite{devries2017dataset} propose interpolating between nearest neighbors of the same class as a way of improving generalization.

Mixup \cite{zhang2017mixup} analyzed the benefits of augmenting the data set with new inputs, on the premises that "linear interpolations of feature vectors should lead to linear interpolations of the associated targets". This can be interpreted as constructing a combinatorial manifold on the feature vectors. Combinatorial manifolds are simply piecewise linear manifolds. This populates the feature space, making adversarial attacks more difficult, while also making the loss landscape more linear, which should lead to a better generalization and faster convergence in first-order optimization frameworks such as gradient descent. Indeed, this approach increased generalization on a variety of tasks, such as image classification, speech recognition, image generation, and others. However, after preliminary tests, they came to the conclusion that combining three or more examples to form a new input does not bring any additional improvements, a result which we will contest in this paper.

SamplePairing \cite{inoue2018japaneseguy} introduces the idea of adding a fine-tuning step, where the original train set is used - without mixing inputs. We also apply this principle and acknowledge its efficacy. In his work, it is shown that smaller data sets benefit from a greater performance gain. It is also claimed that if you don't linearly combine the corresponding targets as well, the performance remains similar, a claim which is not reflected in our results. According to our experiments, not adapting the targets led to a considerable drop in performance.
Another approach by \cite{summers2019improved} of generating outputs by multiple non-linear combinations proved to be just as effective and in some cases better than Mixup \cite{zhang2017mixup} or Between-Class learning \cite{tokozume2018between}.

Mixup has also been extended to work on model hidden states, drastically improving the types of datasets it can work with. The approach presented in \cite{verma2018manifold} finds similar generalization improvements over baseline results as well as robustness to single-step adversarial attacks. An even more interesting approach specializes in improving adversarial defense at a minimal cost to overall clean data accuracy by interpolating clean data with adversarially generated samples \cite{lamb2019interpolated}.

Our contribution consists of a training procedure for generalizing the Mixup \cite{zhang2017mixup}, Between-Class \cite{tokozume2018between} and SamplePairing \cite{inoue2018japaneseguy} approaches so that we can achieve further test accuracy boosts when mixing more than two inputs. This is in contrast to previous works in which it is claimed that using more samples leads to a drop in performance. While we focused our experiments on improving the vanilla mixup version, we believe that our techniques could be applied over other variations, such as the adversarial interpolation \cite{lamb2019interpolated} and hidden state interpolation \cite{verma2018manifold}.
We also propose a test time data augmentation designed to improve accuracy for models trained with mixed data or to be used to increase adversarial attack robustness. This was first described in \cite{pang2019mixup}, but similarly, with Mixup, they were only using interpolations of two samples. As described in this work, we will present a generalized version that is able to make use of more samples when interpolating.

\section{Sum Augmentation} \label{sum_augmentation}
The approach proposed in this paper can be viewed as a generalization of the pairing of input points described by SamplePairing \cite{inoue2018japaneseguy} and MixUp \cite{zhang2017mixup}.
The way a new augmented data sample $x_{new}$ is generated, as described by the previously mentioned papers, is the following:
$$x_{new} = \lambda * x_i + (1- \lambda) * x_j, x_{i,j}\in D \text{ - the original dataset}$$
As described by H. Inoue \cite{inoue2018japaneseguy}, the greatest performance improvement was achieved when $\lambda=0.5$. This result was mirrored in our experiments. In order to generalize the above formula to be applicable to an arbitrary $K$ number of samples, to be linearly combined in the new input, we propose the following:

$$x_{new} = \frac{1}{K} \sum_{i}^{K} x_i, x_{i}\in D \text{ - the original dataset}$$

From the above formulation, it becomes clear that \textbf{Sum Augmentation} increases the domain size of potential inputs from $|D|$ to an estimated:
$$|D_{new}| = {|D|\choose K}$$
This is merely an approximation since one can imagine a dataset D where $\frac{1}{K} \sum_{i}^{K} x_i\in D$, one such dataset could be: $D=\{0,0.5,1\}$, thus $\exists D \text{ where } D \cap D_{new} \neq \emptyset$. Regardless, we believe that in real-life image datasets this overlap would be extremely rare.

In order to extend the above formula to operate on batches in an efficient manner which can be expressed with ease in our implementation we propose the following way to construct a batch of $\lfloor \frac{m}{K} \rfloor$ augmented samples from a batch of $m$ samples of the original dataset:

$$batch_{new}[i] =  \frac{1}{K}  \sum_{j=0}^{K-1} batch[i+j* \left \lfloor{\frac{m}{K}}\right \rfloor] , i\in [1,\left \lfloor{\frac{m}{K}}\right \rfloor]$$ 

From the above definition, our new augmented batch would have $\left \lfloor{\frac{m}{K}}\right \rfloor$ samples, this is why we only chose values for $K$ which were divisors of $m$. To state in a more concise manner, we split the original batch into $K$ groups of equal size, and each element in the augmented batch is computed as the mean of all the elements at the same index in all groups. From now on, we will refer to one such group as a \textit{sum group}.

What we described above is the process undertaken by the input data. From our experiments, we have found that the method which worked best was applying the same linear combination over the one-hot encoding of the labels. As such, the target for our neural network becomes:

$$target_{new}[i]= \frac{1}{K} * c_i$$

Where $c_i$ is the number of samples with the label $i$ which had been summed to obtain the associated augmented input. We chose to calculate the loss by using the binary cross entropy and dividing it by $K$.

\begin{figure}
    \centering
    \caption{$K=2$ Generated Images}
    \includegraphics[height=38px]{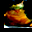}
    \includegraphics[height=38px]{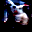}
    \includegraphics[height=38px]{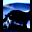}
    \includegraphics[height=38px]{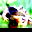}
    \includegraphics[height=38px]{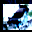}\\
    \includegraphics[height=38px]{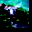}
    \includegraphics[height=38px]{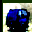}
    \includegraphics[height=38px]{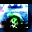}
    \includegraphics[height=38px]{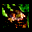}
    \includegraphics[height=38px]{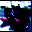}
    \label{fig:$K=2$ Generated Image}

    \caption{$K=4$ Generated Images}
    \includegraphics[height=38px]{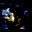}
    \includegraphics[height=38px]{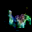}
    \includegraphics[height=38px]{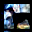}
    \includegraphics[height=38px]{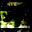}
    \includegraphics[height=38px]{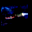}\\
    \includegraphics[height=38px]{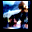}
    \includegraphics[height=38px]{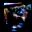}
    \includegraphics[height=38px]{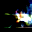}
    \includegraphics[height=38px]{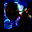}
    \includegraphics[height=38px]{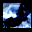}
    \label{fig:$K=4$ Generated Image}

    \caption{$K=8$ Generated Images}
    \includegraphics[height=38px]{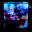}
    \includegraphics[height=38px]{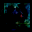}
    \includegraphics[height=38px]{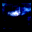}
    \includegraphics[height=38px]{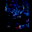}
    \includegraphics[height=38px]{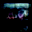}\\
    \includegraphics[height=38px]{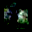}
    \includegraphics[height=38px]{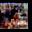}
    \includegraphics[height=38px]{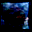}
    \includegraphics[height=38px]{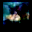}
    \includegraphics[height=38px]{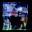}
    \label{fig:$K=8$ Generated Image}
\end{figure}

Examples of these pictures can be seen in figures \ref{fig:$K=2$ Generated Image},\ref{fig:$K=4$ Generated Image} and \ref{fig:$K=8$ Generated Image}. The images still seem to contain shapes and structure, but we found it was a very hard task to visually identify the objects in the images when $K$ is two, and impossible for four or eight. 

Other works have attempted to sample coefficients from a random distribution, such as beta, uniform, or Gaussian. The initial mixup work has achieved better results when doing so, compared to using $\frac{1}{K}$. We weren't able to replicate this behavior in our experiments. While sampling from the beta distribution has proven to be better than using a uniform distribution, both of these methods lagged behind the simple averaging approach.

We did observe that using a beta or uniform distribution to sample coefficients from has lead to a noticeable lower model $L2$ norm, the test accuracy was significantly worse. For example, a WideResNet(28,10) trained on the full CIFAR-10 dataset using the same meta-parameters resulted in a final $L2$ norm of 417.3 for the model trained using beta distribution sampled coefficients. In contrast to a norm of 442.4 when all coefficients were $\frac{1}{K}$.

This drastic difference reflects upon the strong regularisation effects of the technique. This is somewhat an expected behavior since the data domain the model is training on is larger due to the increase in potential inference outputs.

The decrease in norm is usually seen as a positive train since it indicates that the model is not memorizing the training data and overfitting. However, if we regularize a network too strongly, we might end up with an underfit model. When looking at the accuracies, this appears to be the case. The beta distribution sampling version resulted in a 94.51\% accuracy score, which is not only lower than the simple averaging version, but it is weaker than the baseline run result of 95.32\%.

It is possible that different network architectures or datasets which are more prone to overfitting might benefit from the increased norm regularization effect, in which case one might check if sampling from a random distribution offers better results. However, with our current setup of networks and datasets, we chose to go forward with proposing the averaging method since it seems better suited for the task and we believe it presents a better comparison landscape to other similar methods since it removes an additional randomized variable which also requires more meta-parameter fine-tuning.

The approach proposed in this paper can be viewed as a generalization of the simple procedure of pairing input points and their corresponding outputs, resulting in a much larger domain of input points, obtained by linear combinations of the original one.

\section{Cascading Sum Augmentation}
\textbf{Sum Augmentation} works with any number of sum groups. We made use of this fact by training our model in a "cascading" fashion, meaning that we would first train a model using $K$ sum groups, saving a checkpoint for every new best test accuracy. When the model plateaus and doesn't improve anymore we transfer the weights from the best previous checkpoint and continue training using $K/2$ sum groups. The process repeats until $K=1$ when the model does its last fine-tuning step with the original training dataset samples.

\begin{algorithm}[H]
\textbf{procedure $CascadeSumGroups(K)$} \{\\
    Load the previous model checkpoint\\
    \eIf{$K < 1$}{
        \Return
    }{
        \While{loss is decreasing}{
            Update model parameters\\
            Save model checkpoint
        }
        \Return{$CascadeSumGroups(\left \lfloor{\frac{K}{2}}\right \rfloor)$}
    }
\}
 \caption{Cascading sum groups}
\end{algorithm}

Our motivation for this training procedure comes from the observation that while the images resulting from the linear interpolation of 4 or 8 samples is nearly impossible for the human eye to decipher, the model was able to reach validation accuracies way above random chance. For example, a WideResNet(28,10) model trained on the full CIFAR-10 dataset, but only with sum augmentation where $K=8$ reached convergence at a test dataset accuracy of 49.05\%. This result seems incredible when looking at what the model was trained to do, it was given samples of 8 overlapped images and asked to output the proportion of each class that was found in the interpolation. Because of this, we believe that despite the task difficulty and the apparent noisy nature of the data, the model was capable of extracting high-level patterns and general features. Since it seems unlikely that the model could try and memorize small details or overfit on specific class features when the samples it was seeing were sampled from an extremely large potential dataset, where any identifying small class feature would be erased by the overlapping of so many other random samples.

Having this in mind, we wanted to utilize whatever larger level patterns the model was identifying by transferring this knowledge to the easier tasks of interpolating on $\frac{K}{2}$ samples. This process of transferring knowledge from one task to the other being perpetuated until the model trained on clean data. 

\section{Gradually Cascading Sum Augmentation}\label{gradually_cascading_sum_augmentation}
While the approach presented above is able to generate a potential maximum number of samples starting from an initial domain $D$ and a given $K$ of:
$$|D_{total}| = \sum_{j=1}^{K}{|D|\choose j}$$

We foresee that while \textbf{Cascading Sum Augmentation} will be able to successfully transfer the knowledge models gained from higher levels of $K$, a more gradual transition towards training on the original data could potentially produce better results. Our intuition comes from the observation that when we use transfer-learning, the model usually requires an adjustment training period before it can successfully be used on the new data. Since the various levels of $K$ present vastly different classification challenges, in terms of both the input data distribution, but also the expected target results, we believe that a gradual "continuous learning" method should perform better. 

To this end, we have also developed and implemented a training procedure that similarly starts training on interpolations of $K$ images, but will transition to $K-1$ in a smoother, granular way. With the help of Dr. Olariu Florentin Emanuel, we designed a function that will generate interpolation coefficients, which we will use to calculate the augmented data as a coefficient weighted sum of inputs, instead of simply averaging. This function has the desired characteristic of having a control mechanism for granularity. The \textbf{Gradually Cascading Coefficient (GCC)} function is calculated using the following components:
\begin{gather}
    k = \lfloor t \cdot (n - 1) \rfloor \\
    centroid = \frac{1}{n-k} + \frac{t \cdot (n-1) - k}{(n-k) \cdot (n-k-1)} \\
    remainer = 1 - (centroid \cdot (n-k-1)) \\
    eps = remainder \cdot \frac{t-\frac{k}{n-1}}{n-1}
\end{gather}
Where:
\begin{itemize}
    \item $n$ is the starting number of inputs we wish to interpolate, it can also be seen as the initial value of $K$.
    \item $t$ is a granularity control variable, it is defined as $t\in [0,1]$. Intuitively, $t$ represents the percentage of how much of the transition from $K$ to $1$ we have completed.
\end{itemize}
The GCC function can then be defined as:
\begin{gather}
    GCC_n \colon [0,1] \to [0,1]^n \\
    GCC_n(t) = (\underbrace{centroid +\frac{eps}{n-k-1}, centroid + \frac{eps}{n-k-1} \dots}_{n-k-1}, remainer-eps, \underbrace{0, 0\dots}_{k})
\end{gather}
The function does not present itself to be very intuitive, and it is clearly an artificial construct. This is because we took a bottom-up approach in modeling the function, starting from desired output examples, and building rules to satisfy our constraints. However, upon closer consideration, some intuitive explanations can be extracted: 
\begin{itemize}
    \item As obvious from the formulations, $k$ represents how many inputs we have eliminated from the initial configuration. Because of this, we have: $K=n-k$ for the current value of $K$.
    \item The $centroid$ variable represents the weight which would be used if we desired to calculate the average of $n-k$ samples. It can be seen that when $k = t\cdot(n-1)$, $centroid = \frac{1}{n-k}$.
    \item The values at positions 1,2,3\dots $n-k-1$ represent the \textit{"main"} coefficients, while the value at position $n-k$ will be the coefficient that is currently being gradually reduced towards 0. The remaining positions are coefficients which have already been set to 0.
    \item As such, calculating a weighted sum using these coefficients will effectively calculate an interpolation of the first $n-k$ samples. The sample at position $n-k$ is gradually reduced as $t$ gradually increases until it reaches 0, at which point $k$ increases by 1 and we start using one less sample. This process repeat continues until we reach $n-k=1$ when we are using non-augmented data.
    \item The $remainer$ simply represents how much \textit{"unallocated"} percentage is left to be distributed to coefficients. It can be seen as how much would be left on the $n-k$ position, such that the resulting coefficients sum to $1.0$.
    \item We use $eps$ as a construct with which we can distribute part of the value of the \textbf{"remainer"} coefficient to the \textbf{"main"} coefficients. This way the centroid values are moving from $\frac{1}{n-k}$ closer to $\frac{1}{n-k-1}$, while the decreasing coefficient is going to 0. This happens proportionally to the values taken by $t$ in $t\in (\frac{k}{n-1},\frac{k+1}{n-1}]$. As stated previously $t$ is the variable which controls what \textbf{"percentage"} of the coefficient interval we have covered.
\end{itemize}

Since we are going to calculate the weighted sum of data inputs instead of averaging, the process of constructing an augmented batch and target must be modified to the following form:
\begin{gather*}
    C = GCC_n(t) \\
    batch_{new}[i] =  \frac{1}{C[i]}  \sum_{j=0}^{K-1} batch[i+j* \left \lfloor{\frac{m}{K}}\right \rfloor] , i\in [1,\left \lfloor{\frac{m}{K}}\right \rfloor] \\
    target_{new}[i]= \frac{1}{C[i]} * c_i
\end{gather*}

Using the above formulation, we propose the following \textbf{Gradually Cascading Sum Augmentation} algorithm:

\begin{algorithm}[H]
\textbf{procedure $GraduallyCascadingSumGroups(K)$} \{\\    
    $n \gets K$ \\
    $t \gets 0.0$ \\
    $t_{step} \gets 1.0/nr\_epochs$ \\
    \For{epoch in nr\_epochs}{
        Update model parameters \\
        $t \gets t + t_{step}$
    }
    \For{epoch in nr\_finetune\_epochs}{
        Update model parameters
    }
\}
\caption{Gradually cascading sum groups}
\end{algorithm}

The above algorithm dynamically adjusts the granularity of the cascade to fit in the given number of epochs. With this, the variable $t$ starts from 0 and finishes reach 1 exactly at the last epoch, after which, the algorithm would be training on clean data. Similarly to the previous version, we introduce a fine-tuning step to our training process, which will train the network using normal images (since $t=1$) for a given number of epochs.

As can be seen, the only modifications required to the classic ANN training algorithm are the addition of a processing step for data augmentation and a few lines of code to initialize and increment $t$. This way, our proposed method should be compatible with any other training techniques or datasets for which data interpolation approaches can be used.

\begin{figure}[ht]
    \includegraphics[width=1.0\textwidth]{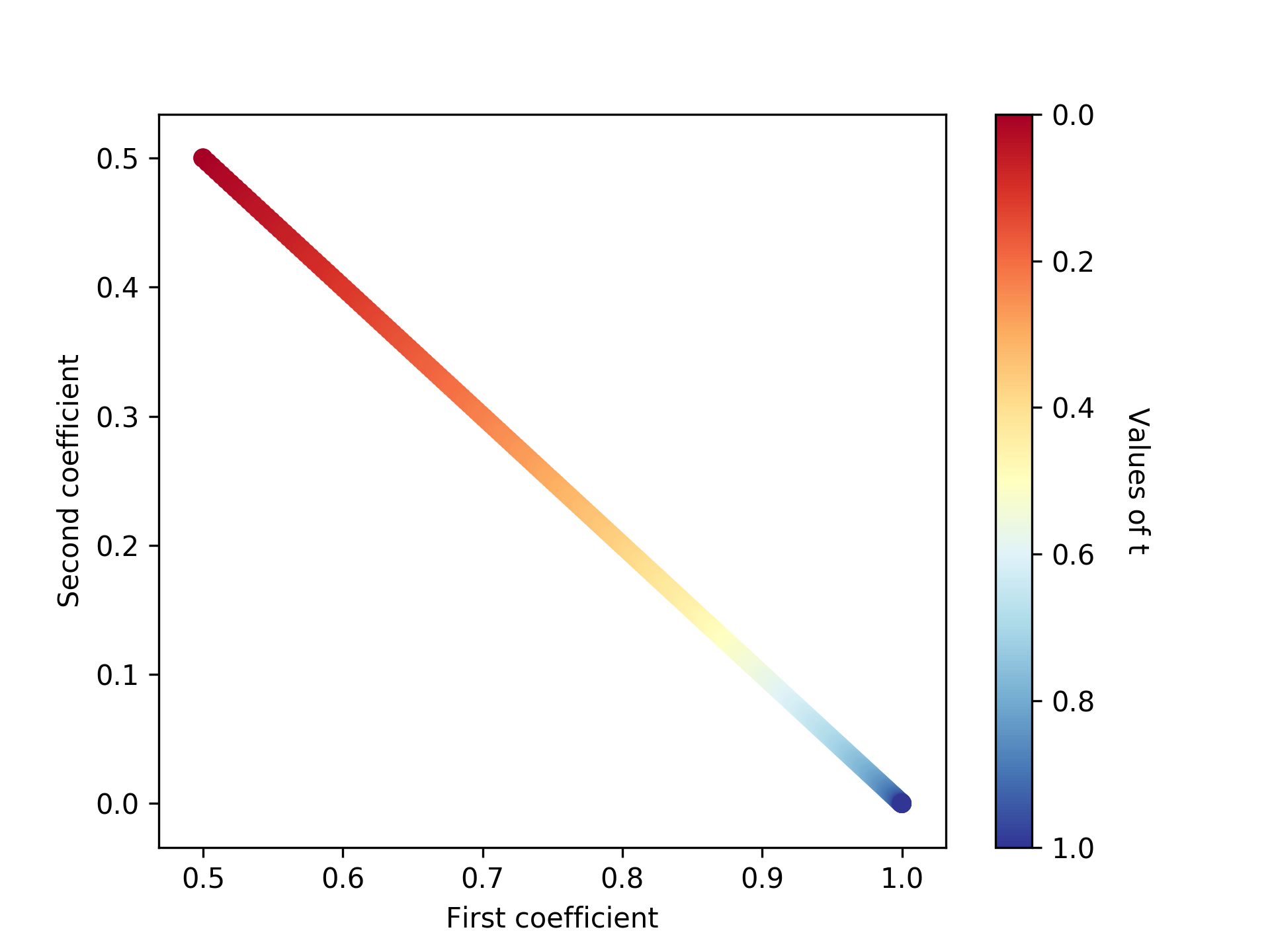}
    \caption{$GCC_2(t)$ function plot}
    \label{fig:gcc2d}
\end{figure}

The $GCC_2(t)$ function plot can be seen in Figure \ref{fig:gcc2d}. Similar behavior would be observed for higher dimensions but those become too difficult to visualize. The expected behavior is that one coefficient is always reducing and has its value distributed to the other coefficients. After it reaches 0, it is ignored and we jump to the next coefficient to reduce to 0 until we are left with only one coefficient with a value of 1.

An important fact to consider is that our current formulation was applied to linear interpolations. We chose this because the initial mixup work and most subsequent papers have utilized linear interpolations. While some results were obtained using non-linear interpolations, this remains the most used type of mixup technique.

\section{Test Time Sum Augmentation} \label{test_time}
Deep neural networks are prone to form non-robust features \cite{ilyas2019adversarial}, which have proven essential to good performance but are prone to adversarial attacks \cite{szegedy2013intriguing} and limit the potential accuracy of the model. This happens because the model has overfitted to classify certain features as a certain class and images containing even a small part of such a feature may be easily misclassified. This is the fault that is being exploited in many adversarial attacks, such as the one described by Su et al. in One pixel attack \cite{su2019onepixel}.

Considering this and taking inspiration from \textbf{Sum Augmentation}, we propose a generalized Mixup Inference \cite{pang2019mixup} test time augmentation method which boosts the accuracy of models trained with $K>1$ sum augmentation and that can be used to mitigate adversarial attacks.

Assume we want to test the predictions of a trained model on a given image $B_i$ with target $Target_i$. This procedure takes a parameter $C$, and performs the following operations:

\begin{enumerate}
    \item Generate $C$ \textit{"augmented"} images, where each image is the linear combination, of the original image and a new random image from the test dataset ($RandomSample()$ returns a random image from the train or test dataset)
    $$ Aug_i = \lambda \frac{1}{K}B_i + \sum_{j=0}^{K-2}\frac{1}{K\lambda} RandomSample(), \lambda\in [0,1] $$
    We consider $\lambda$ to be a parameter that determines how much weight should the images we want to infer on have in the resulting augmented input. This is useful as a mechanism for controlling a trade-off between additional robustness and clean image prediction performance. A $\lambda = 2.0$ would mean we want to give two times the importance to the sample we actually desire to infer upon. For the remainder of the article, we will only be considering $\lambda=1.0$ such that the resulting linear interpolation will effectively average the inputs.
    \item Use the network to generate predictions for each of the augmented images (resulting in $C$ 1D tensors of size $num\_classes$):
    
    $$ Pred_i = predict(Aug_i) $$
    
    \item Compute the output of the network for each image as the mean of all predictions:
    
    $$ Out = (Pred_1 + Pred_2 + ... + Pred_c) / C $$
    
    \item At this point, $Out$ is a 1D tensor of size $num\_classes$. So we take the $argmax$ of the output of the network over all classes and compare it to the corresponding $Target_i$ and compute the results (accuracy, loss, F-score, or others).
\end{enumerate}

The motivation behind this method lies in the fact that a given sample $x_i$ will be forwarded through the network multiple times, each time summed with one or more different random samples. This should have the effect of obtaining, on average, a higher activation for the correct label than the other ones. The benefit of using this \textbf{Test Time Sum Augmentation} is that it limits the effect that over-fitted features have in classifying a given image. \\

Preliminary experiments showed that this makes the model much more \textbf{robust against adversarial attacks}. We hypothesize that this is happening because it is significantly more difficult to generate noise that is effective for all the augmented images created. By computing the average vote over all $C$ augmented images classified by the model, we are forcing the adversary to generate an example which will be summed with $C$ other randomly chosen samples, these resulted augmented inputs have to, on average, produce a classification error in the majority of the augmented data points. \\

\section{Experiments} \label{experiments}

In order to showcase the improvement our generalized method provides, we use the CIFAR-10 and CIFAR-100 datasets. Data augmentation techniques artificially increase the size of the dataset we have available. This makes them bring the biggest boost in performance, especially when applied to small datasets. In order to illustrate this, we created subsets of different sizes of our datasets to emulate a scenario where we would only have access to a very small number of samples. This allowed us to observe what are the expected boosts in performance our method can bring. The results should be of interest, especially for medical datasets, where the acquisition of additional data is costly or simply unfeasible. The following subset sizes were used:
\begin{itemize}
    \item CIFAR-10
    \begin{itemize}
        \item 1000 samples (100 samples per class)
        \item 5000 samples (500 samples per class)
        \item 50000 samples (the original dataset)
    \end{itemize}
    \item CIFAR-100
    \begin{itemize}
        \item 10000 samples (100 samples per class)
        \item 50000 samples (the original dataset)
    \end{itemize}
\end{itemize}

We considered that experimenting on both CIFAR datasets was important in order to show that increasing the number of classes does not affect the boost in performance gained when using our method.

Since we wanted to see if our method is still able to produce noticeable boosts in performance when scaling the size of the model, we chose to train two popular WideResNet architectures \cite{zagoruyko2016wideresnet} of different sizes:
\begin{itemize}
    \item  The 40 layers deep model with a widening factor of 4 having 8.9M parameters
    \item  The 28 layers deep model with a widening factor of 10 having 36.5M parameters
\end{itemize}

All models have been trained with a batch size of $100$, and a learning rate of $0.05$. When the learning stagnates for 300 gradient update steps, the learning rate is reduced by a factor of $0.5$. The optimizer we used was Stochastic Gradient Descent \cite{ruder2016sgdoverview} with a Nesterov accelerated momentum \cite{sutskever2013initialization} of $0.9$. The $L2$ penalty for regularizing the model was set to $0.0001$. We have chosen to use the same parameters across all our runs in order to achieve a fair comparison between them. This means that it is very likely that the best result achievable by \textbf{Sum Augmentation} could be even higher than the one presented below since no optimal meta-parameter search has been attempted for each specific configuration. While any experiment configuration would have potentially benefited greatly by meta-parameter, tuning it would have reduced the clarity of what are the exact effects of using our technique.

\subsection{Cascading Sum Augmentation Results}

\begin{table*}[ht]
    \caption{CIFAR-10 Test Error using Cascading Sum Augmentation} \label{tab:CIFAR-10 Test Error using Cascading Sum Augmentation}
    \begin{center}
    \begin{tabular}{l c c c | c c c c}
        \cmidrule[2pt]{1-7} 
        \multirow{2}{*}{\textbf{Starting Sum Groups}} & \multicolumn{3}{c|}{WideResNet(40,4)} & \multicolumn{3}{c}{WideResNet(28,10)} \\ 
        \cmidrule{2-7}
        & 100 & 500 & 5000 & 100 & 500 & 5000 & (Samples/Class) \\ 
        \cmidrule{1-7}
        \hfill 8 & 31.19\% & 13.51\% & 4.02\%  & \textbf{29.34\%}  & 13.21\%  & 3.28\%  \\
        
        \hfill 4 & \textbf{30.65\%} & \textbf{13.29\%} & \textbf{3.88\%}  & 30.36\%  & \textbf{12.81\%}  & \textbf{3.11\%}  \\
        
        \hfill 2 & 32.44\% & 14.88\% & 5.53\%  & 30.99\%  & 13.1\%  & 3.44\%  \\
        
        \hfill Baseline & 47.36\% & 20.15\% & 5.57\%  & 43.47\%  & 19.96\%  & 4.68\%  \\
        \cmidrule[2pt]{1-7}
    \end{tabular}
    \end{center}
\end{table*}

\begin{table*}[ht]
    \caption{CIFAR-100 Test Error using Cascading Sum Augmentation} \label{tab:CIFAR-100 Test Error using Cascading Sum Augmentation}
    \begin{center}
        \begin{tabular}{l c c | c c c}
            \cmidrule[2pt]{1-5} 
            \multirow{2}{*}{\textbf{Starting Sum Groups}} & \multicolumn{2}{c|}{WideResNet(40,4)} & \multicolumn{2}{c}{WideResNet(28,10)} \\ 
            \cmidrule{2-5}
            & 100 & 500 & 100 & 500 & (Samples/Class) \\ 
            \cmidrule{1-5}
            \hfill 8 & 35.3\% & 20.34\% & 35.01\% & 18.63\% \\
            
            \hfill 4 & \textbf{34.7\%} & \textbf{19.85\%} & \textbf{33.01\%} & \textbf{18.09\%} \\
            
            \hfill 2 & 35.56\% & 20.17\% & 34.44\% & 18.17\% \\
            
            \hfill Baseline & 43.2\% & 23.87\% & 41.72\% & 21.94\% \\
            \cmidrule[2pt]{1-5}
        \end{tabular}
    \end{center}
\end{table*}

\begin{figure}[ht]
    \includegraphics[width=\textwidth]{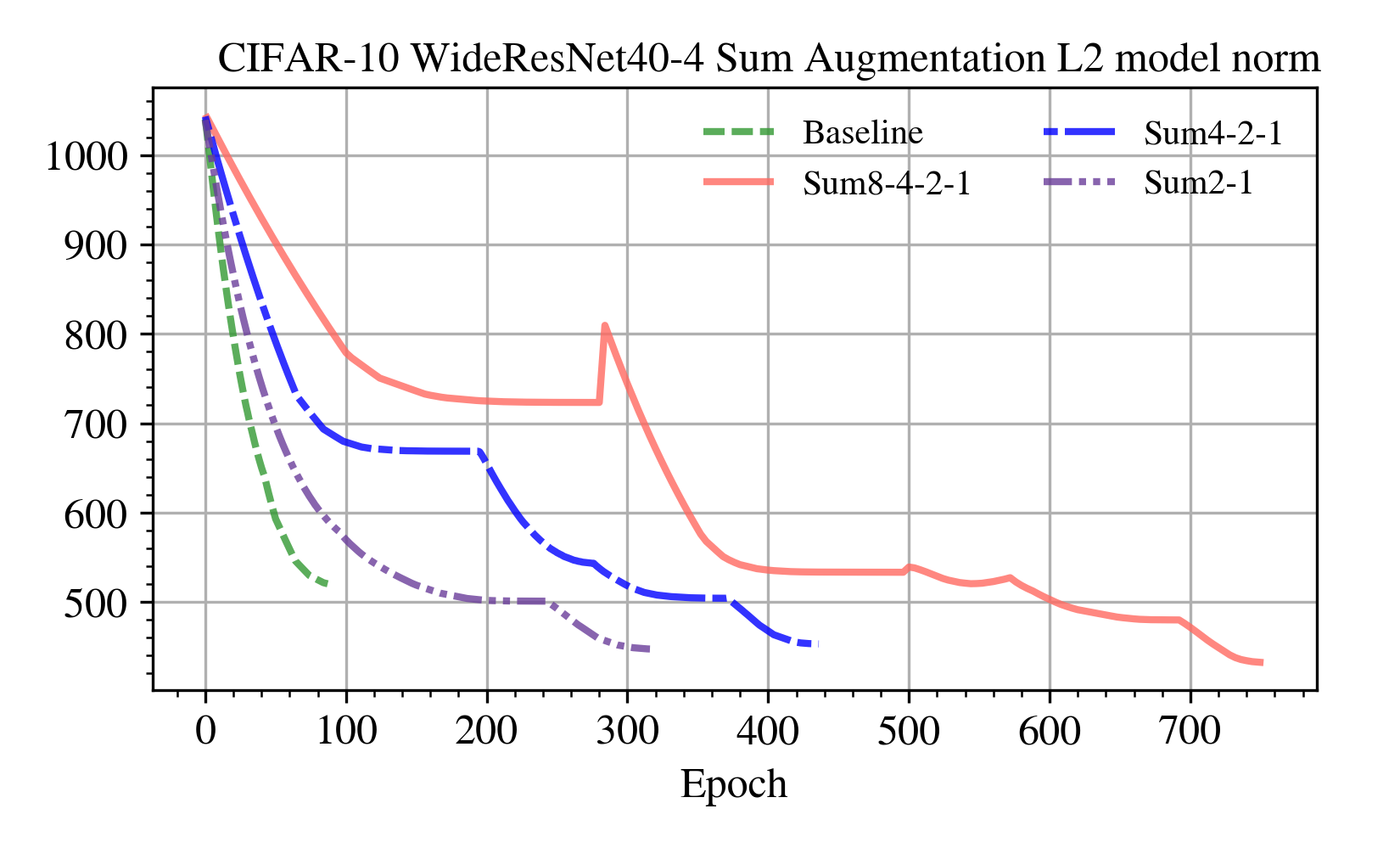}
    \caption{$L2$ norm}
    \label{fig:l2_norm}
\end{figure}

An easily observable property is that using \textbf{Cascading Sum Augmentation} with higher values for $K$ leads to a lower $L2$ norm of the final model (Figure \ref{fig:l2_norm}). This further emphasizes the regularisation effect of \textbf{Sum Augmentation}, which previous works have observed. This effect is to be expected since models trained with the cascading method are able to train for more epochs before they plateau and stop learning. We notice that there are spikes in the value of the norm at the points where the model switches from $K$ to $\frac{K}{2}$ sum groups. This hints towards the possibility of a future improvement by having smoother cascade transitions, for example, taking steps from $K$ to $K-1$ when the model stops learning, instead of having it. The loss values through training follow a similar behavior.

An interesting question is whether higher-order Sum Groups ($K=8$ for example) have no useful features left that the model could learn, since averaging over so many samples produces, what seems to be, a very "noisy" image. Rather surprisingly, the accuracy turns out to be much higher than expected. On CIFAR-10 using the WideResNet(28,10) model, Sum Augmentation with $K=8$ reached an accuracy score of $49.05\%$, being trained only on images like the ones seen in Figure \ref{fig:$K=8$ Generated Image}.

We found that this \textbf{Cascading Sum Augmentation} method outperforms training with only two groups (k=2) and fine-tuning using the original data (as proposed by Zhang et al. \cite{zhang2017mixup} and Inoue et al. \cite{inoue2018japaneseguy}) in all the cases we tested, by a significant margin.

As it can be seen in tables \ref{tab:CIFAR-10 Test Error using Cascading Sum Augmentation} and \ref{tab:CIFAR-100 Test Error using Cascading Sum Augmentation}, when data is very limited, the model with higher learning capacity, $WideResNet(28,10)$, achieves the lowest error rate. This can be seen when starting the Cascading algorithm with $K=8$ Sum Groups, trained on a subset of CIFAR-100, containing only 100 samples per class. One way to interpret this is that due to the higher learning capacity, the model can still identify useful patterns in the data, even if it receives 8 overlapped images as input. When more data is available or the model is smaller, $K=4$ seems to consistently have better performance. We noted the run where $K=1$ as being our "Baseline" since it is trained using the normal dataset, not making use of \textbf{Sum Augmentation} and it is the run which we will calculate the increase in accuracy from. 

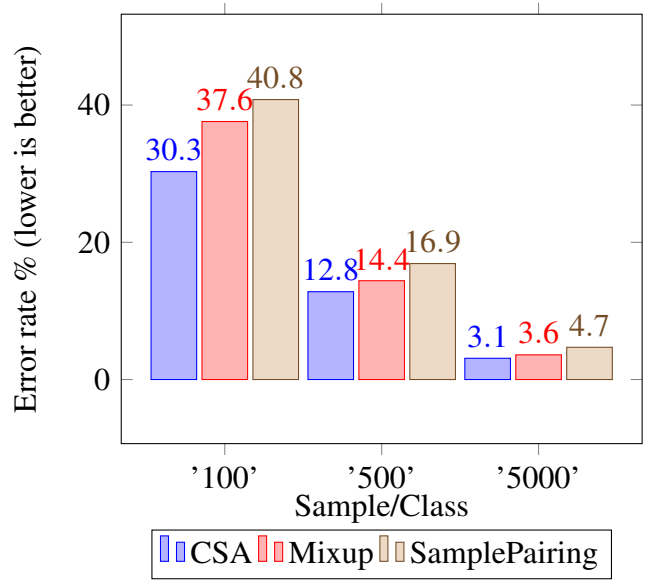
\begin{figure}
    \centering
    \begin{tikzpicture}
        \begin{axis}[
            ybar,
            bar width=0.6cm,
            enlargelimits=0.33,
            legend style={at={(0.5,-0.20)},
              anchor=north,legend columns=-1},
            xlabel={Sample/Class},
            ylabel={Error rate \% (lower is better)},
            symbolic x coords={'100','500','5000'},
            xtick=data,
            nodes near coords,
            nodes near coords align={vertical},
            ]
        \addplot coordinates {('100',30.3) ('500',12.8) ('5000',3.1)};
        \addplot coordinates {('100',37.6) ('500',14.4) ('5000',3.6)};
        \addplot coordinates {('100',40.8) ('500',16.9) ('5000',4.7)};
        \legend{CSA,Mixup,SamplePairing}
        \end{axis}
    \end{tikzpicture}
    \caption{Comparison on CIFAR-10}
    \label{fig:other_methods_cifar10}
\end{figure}

In order to compare our method with similar methods, we ran the experiments described in tables \ref{tab:CIFAR-10 Test Error using Cascading Sum Augmentation} and \ref{tab:CIFAR-100 Test Error using Cascading Sum Augmentation} using the Mixup \cite{zhang2017mixup} and SamplePairing \cite{inoue2018japaneseguy} methods. In Figure \ref{fig:other_methods_cifar10} you can see the comparison between our method CSA (Cascading Sum Augmentation) with $K=4$ (the most consistent parameter value for the given task) on CIFAR-10 using WideResNet(28,10). Our method's proficiency in small datasets is even clearer when comparing the accuracy boosts, each method bringing over the baseline performance. Our method brings a 30.15\% error reduction while Mixup, the second-best performing method, only decreases the error rate by 13.5\%. This gap is smaller, but still considerable when training on the entire dataset, our method producing a 33.54\% decrease in error while Mixup achieves only a 22.86\% reduction.

Another interesting finding, which matches the results of previous work, is the increase in model robustness to adversarial attacks. It is unclear how much of the boost is due to the nature of our technique and how much is simply the result of the greater generalization power that the model reaches. This can be seen in table \ref{tab:CIFAR-10 Adversarial Defence Comparison}.
\begin{table}
    \caption{CIFAR-10 Adversarial Defence Accuracy Comparison} \label{tab:CIFAR-10 Adversarial Defence Comparison}
    \begin{center}
    \begin{tabular}{l c c c c c}
        \cmidrule[2pt]{1-6}
        & Clean & FGSM\cite{goodfellow2014explaining} & BIM\cite{kurakin2016adversarial} & MIN\cite{dong2018boosting} & PGD\cite{madry2017towards} \\ 
        \cmidrule{1-6}
        \hfill Baseline & 95.32\% & 24.08\% & 0.2\% & 0.0\% & 0.0\% \\
        \cmidrule{1-6}
        \hfill \textbf{CSA} & \textbf{94.38\%} & \textbf{54.0\%} & \textbf{17.2\%} & \textbf{9.19\%} & \textbf{3.22\%} \\
        \cmidrule[2pt]{1-6}
    \end{tabular}
    \end{center}
\end{table}

The adversarial attacks we chose are all gradient-based attacks, which have a devastating effect on model accuracy. While arguably our method does not offer a practically sufficient boost in robustness, we must accentuate the fact that this result was obtained on a network that did not benefit from any other adversarial defense technique during training.

\subsection{Gradually Cascading Sum Augmentation Results}
As indicated by the spiking behavior of the model norm, as well as its loss, sudden jumps from an input data domain, which interpolates 8 images to a domain, which interpolates only 4, introduce a period of training where the model has to adjust to the new input domain. This characteristic is expected and generally found when executing transfer learning between different datasets.

\begin{table}
    \caption{CIFAR-10 Test Error using Gradually Cascading Sum Augmentation} \label{tab:CIFAR-10 Test Error using Gradually Cascading Sum Augmentation}
    \begin{center}
    \begin{tabular}{l c c c}
        \cmidrule[2pt]{1-3} 
        \multirow{1}{*}{\textbf{Starting Sum Groups}} & \multicolumn{1}{c}{WideResNet(40,4)} & \multicolumn{1}{c}{WideResNet(28,10)} \\ 
        \cmidrule{1-3}
        \hfill 8 & 3.89\% & 3.17\%  \\
        
        \hfill 4 & \textbf{3.71\%} & \textbf{2.95\%}  \\
        
        \hfill 2 & 5.25\% & 3.27\%  \\
        
        \hfill Baseline & 5.57\% & 4.68\%  \\
        \cmidrule[2pt]{1-3}
    \end{tabular}
    \end{center}
\end{table}

\begin{figure}[ht]
    \includegraphics[width=\textwidth]{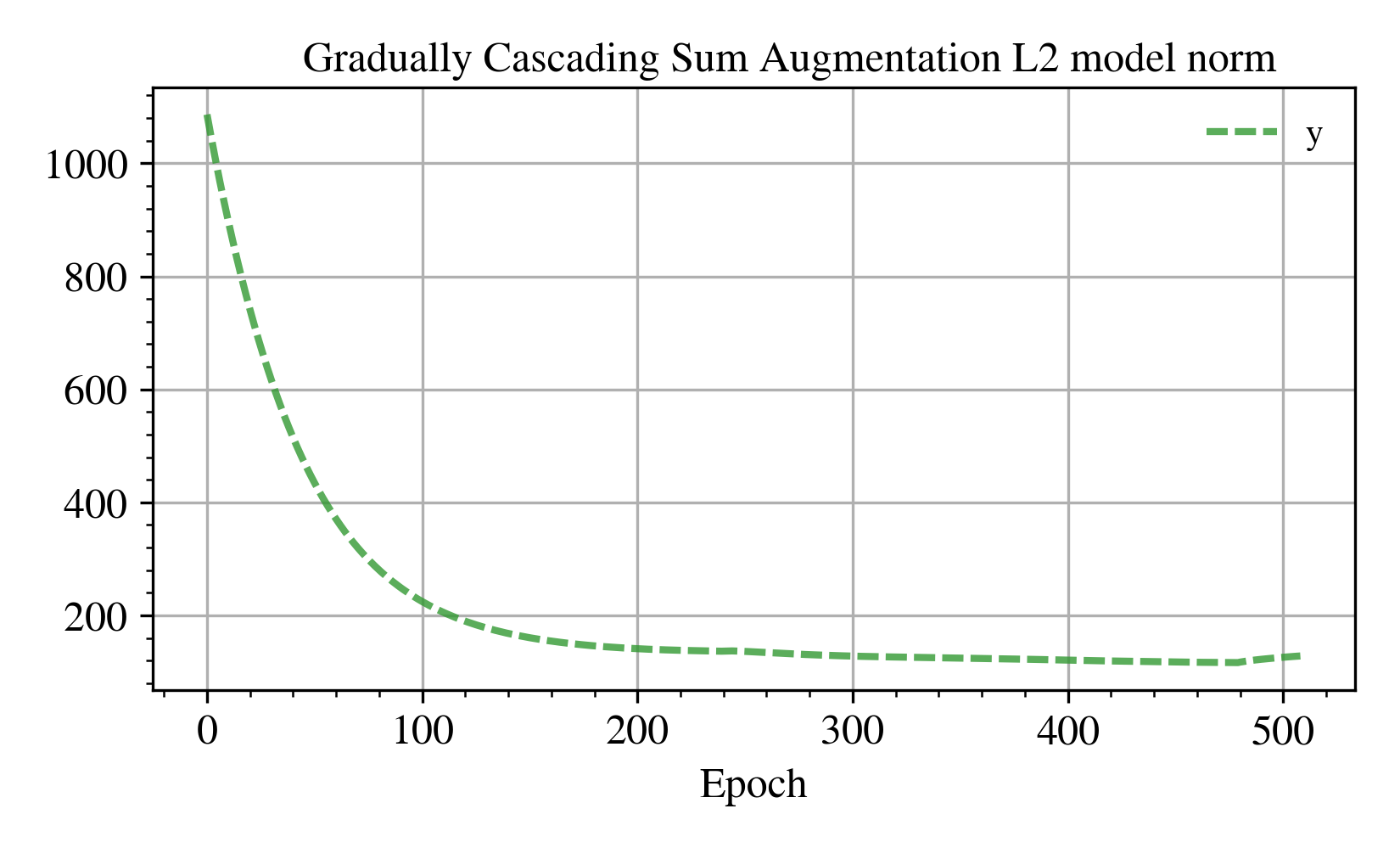}
    \caption{Gradually Cascading $L2$ norm}
    \label{fig:gradual_l2_norm}
\end{figure}

Our proposed \textbf{Gradually Cascading Sum Augmentation} training procedure was designed to alleviate that issue. As it can be seen from Figure \ref{fig:gradual_l2_norm}, this seems to be the case, the norm and loss of the model no longer spike in values, which required additional training epochs in order to readjust.

As it can be seen in table \ref{tab:CIFAR-10 Test Error using Gradually Cascading Sum Augmentation}, this method achieves around a 5\% performance boost over \textbf{Cascading Sum Augmentation}. 

\subsection{Test Time Sum Augmentation Results}

\begin{table*}[ht]
    \caption{CIFAR-10 Test Error using Test Time Sum Augmentation} \label{tab:CIFAR-10 Test Error using Test Time Sum Augmentation}
    \begin{center}
    \begin{tabular}{l c c | c c}
        \cmidrule[2pt]{1-5} 
        \multirow{2}{*}{\textbf{$K$ values}} & \multicolumn{2}{c|}{WideResNet(40,4)} & \multicolumn{2}{c}{WideResNet(28,10)} \\ 
        \cmidrule{2-5}
        & Normal Test & Test Time Sum Aug & Normal Test & Test Time Sum Aug\\ 
        \cmidrule{1-5}
        \hfill 8-4-2 & 6.53\% & 6.32\%  & 5.58\%  & 5.11\% \\
        \hfill 4-2 & 6.91\% & 6.32\%  & 5.35\%  & \textbf{5.09\%} \\
        \hfill 2 & 6.19\% & \textbf{6.03\%}  & 5.74\%  & 5.39\% \\
        \cmidrule[2pt]{1-5}
    \end{tabular}
    \end{center}
\end{table*}

\begin{table*}[ht]
    \caption{CIFAR-100 Test Error using Test Time Sum Augmentation} \label{tab:CIFAR-100 Test Error using Test Time Sum Augmentation}
    \begin{center}
    \begin{tabular}{l c c | c c}
        \cmidrule[2pt]{1-5} 
        \multirow{2}{*}{\textbf{$K$ values}} & \multicolumn{2}{c|}{WideResNet(40,4)} & \multicolumn{2}{c}{WideResNet(28,10)} \\ 
        \cmidrule{2-5}
        & Normal Test & Test Time Sum Aug & Normal Test & Test Time Sum Aug\\ 
        \cmidrule{1-5}
        \hfill 8-4-2 & 27.18\%  & 24.06\% & 25.11\% & \textbf{20.25\%} \\
        \hfill 4-2 & 27.22\%  & 23.9\% & 24.42\% & 20.46\% \\
        \hfill 2 & 27.19\% & \textbf{23.58\%} & 24.32\% & 20.29\% \\
        \cmidrule[2pt]{1-5}
    \end{tabular}
    \end{center}
\end{table*}

The results of the experiments using \textbf{Test Time Sum Augmentation} can be found in tables \ref{tab:CIFAR-10 Test Error using Test Time Sum Augmentation} and \ref{tab:CIFAR-100 Test Error using Test Time Sum Augmentation}. These results were obtained by running our test time data augmentation method on the best-performing weights of models trained with the cascading algorithm with one small alteration. For this method, we chose to stop the cascading algorithm when reaching $K=2$. The reason for this choice is that all runs that reached $K=1$ would have worse performance when inferring on data using \textbf{Test Time Sum Augmentation}. This is to be expected since when a model would fine-tune on the normal dataset it would have to adjust to the new sample distribution, hence, leading to some degree of \textit{Catastrophic forgetting} \cite{Pf_lb_2018}.

As it can be seen from the test error rates, our test time data augmentation method leads to considerable boosts in prediction performance for models whose final fine-tuning is done on \textbf{Sum Augmentation} with $K=2$. Comparing these results with the ones in tables \ref{tab:CIFAR-10 Test Error using Cascading Sum Augmentation} and \ref{tab:CIFAR-100 Test Error using Cascading Sum Augmentation} it is clear that the method is not suitable for being used when we only want to obtain a model with the best test data accuracy.

We consider the use-case where \textbf{Test Time Sum Augmentation} should be applied is when we are either required or desire to utilize models trained on linear combinations of samples. As stated in section \ref{test_time}, from limited preliminary experiments, our proposed test time data augmentation was shown capable of improving the robustness of a deep model to adversarial examples. 

Our limited preliminary experiments consisted of applying \textbf{Test Time Sum Augmentation} over the baseline WideResNet(28,10) model trained on the normal CIFAR-10 dataset and the meta-parameters described earlier. We chose to test the baseline results instead of one of our technique results since we wanted to show if this test time augmentation has general applicability and not just an edge case scenario application. These results can be seen in table \ref{tab:CIFAR-10 Adversarial Defence Comparison}. 
\begin{table}
    \caption{CIFAR-10 Adversarial Defence Accuracy Comparison} \label{tab:CIFAR-10Adversarial Defence Comparison}
    \begin{center}
    \begin{tabular}{l c c c c c}
        \cmidrule[2pt]{1-6}
        & Clean & FGSM & BIM & MIN & PGD \\ 
        \cmidrule{1-6}
        \hfill Baseline & 95.32\% & 24.08\% & 0.2\% & 0.0\% & 0.0\% \\
        \cmidrule{1-6}
        \hfill Sum-Aug & \textbf{94.38\%} & \textbf{54.0\%} & \textbf{17.2\%} & \textbf{9.19\%} & \textbf{3.22\%} \\
        \cmidrule[2pt]{1-6}
    \end{tabular}
    \end{center}
\end{table}

The adversarial attacks we chose are all gradient-based attacks, which have a devastating effect on model accuracy. While arguably our method does not offer a practically sufficient boost in robustness, we must accentuate the fact that this result was obtained on a network that did not benefit from any other adversarial defense technique during training. It is also important to point out that this test time augmentation could feasibly be implemented alongside any number of other test-time adversarial defense methods and as it can be seen, the drop of generalization on clean data is relatively small compared to the potential gains. And this trade-off between robustness and generalization on clean data, can be adjusted by simply modifying the $\lambda$ parameter, in this test we used $\lambda = 1.0$, $K = 4$ and $C = 16$.

\section{Conclusion} \label{conclusion}
We presented a generalized mixup approach data augmentation procedure that outperforms previous similar work on CIFAR-10 and CIFAR-100, which could be integrated into any Deep Learning algorithm working on these datasets and producing a significant generalization performance gain. The improvements can be as great as a 33.54\% decrease in test data error rate without doing an extensive meta-parameter search.

An immediate potential improvement for our cascading procedure would be to have a more gradual drop in the number of samples that are being mixed to obtain augmented data. We propose an algorithm to achieve this, called \textbf{Gradually Cascading Sum Augmentation}. This causes the \textbf{Sum Augmentation} algorithm to gradually decrease the importance each sample has in forming the mixed data, reaching a point where the model is effectively training on the original dataset. Since this is done gradually with a controllable granularity mechanism, we have control over how sudden the input distribution of our model changes.

Further work is required to investigate if these results can be replicated on additional object classification datasets or even extended on different computer vision tasks, such as object detection and image segmentation. Additional research is required to determine to what extent \textbf{Cascading Sum Augmentation} produces models that are resistant to adversarial attacks compared to related work. If any positive result is reached, an analysis would be required to determine if the increase in resistance to attacks is due only to the improved test data generalization (already reflected in the accuracy) or if these models obtain an additional intrinsic robustness property.

Another interesting procedure with further research potential being the \textbf{Test Time Sum Augmentation} technique. Having the configurable $\lambda$ parameter, one could experiment with its value as being a trade-off between performance and robustness. Higher values correlate with "cleaner" test data to be classified for better performance, while still keeping a degree of increase robustness, while smaller values should lead to a more resilient model. A more extensive comparison to Mixup Inference \cite{pang2019mixup} is required in order to determine whether the generalization of being able to use more samples per forwarding pass improves robustness or decreases the number of iterations required.

%% file: Conclusions/conclusions.tex
\pagenumbering{gobble}
\begin{conclusions}
We have presented two training procedure augmentations meant to either improve previous work results by extending them such as we did with \textbf{Cascading Sum Augmentation} for mixup methods. The other contribution focused more on generalising a known framework for structuring training data for increased convergence speed. We presented a more general framework to develop data structuring strategies and we also show off a few example strategies, some of which are valid candidates to be used as warm-up methods for neural networks. More interestingly we hoped to show more emphasis on the potential speed and generalization improvements by smartly ordering the training data.

In addition to the proposed future work presented through the paper, we have the following list of future projects:
\begin{enumerate}
    \item \textbf{Loss Standard Deviation Regularisation}
    
    \textbf{Basic idea:} Add a regularization parameter computed from the standard deviation of loss values of samples within a batch.
    
    The idea behind this being that if a model has drastically different loss values for given samples, it would indicate the network memorized features for some of the inputs. This regularization is meant to punish that.
    \item \textbf{Residual Momentum}
    
    \textbf{Basic idea:} Use a sliding window or moving average on residual values to speed up training.
    
    Partial positive results were obtained using a moving average but, it often leads to exploding values, a more stable method such as a sliding window needs to be experimented with.
    \item \textbf{Random Feature-Map Augmentation}
    
    \textbf{Basic idea:} Use data augmentation techniques at a latent representation level in models. The first experiments should look at convolutional networks and apply these transformations on feature maps since they conserve a similar representation to the original data.
    \item \textbf{Gradual Top-K Attention}
    
    \textbf{Basic idea:} Using a similar approach to \textbf{Gradually Cascading Sum Augmentation}, each layer has a set of modules, and for each input, they will be weighted with their attention score. After that choose only top n-1 for weighted sum and then top n-2
    \item \textbf{Embedding Target for NLP}
    
    \textbf{Basic idea:} Instead of using a one-hot output for NLP tasks, output an embedding. Calculate the distance between the output embedding to the embeddings of the targets, the distances will compose a vector which can then simply be softmax and used as the final one-hot output of the model.
    \item \textbf{Overfit Fix via Sample Embedding}
    
    \textbf{Basic idea:} Assign an embedding to each sample. Propagate the gradient to each sample. After you are done learning, either slowly penalize the norm of each embedding, either slowly reduce the norm of each embedding by dividing by a specific fraction of its norm.
\end{enumerate}
\end{conclusions}